\RequirePackage{fix-cm}
\documentclass[twocolumn]{svjour3}          
\smartqed  
\usepackage{epsfig}
\usepackage{graphicx}
\usepackage{amsmath}
\usepackage{amssymb}
\usepackage{bm}
\usepackage{subfigure}
\usepackage{algorithm}
\usepackage[update,prepend]{epstopdf} 
\usepackage{hyperref}
\usepackage[square,comma,numbers,sort&compress]{natbib}
\usepackage{multirow}
\usepackage{color,colortbl}
\definecolor{Gray}{gray}{0.95}

\usepackage[switch]{lineno} 

\journalname{IJCV}
\usepackage{xspace}
\usepackage{eso-pic}

\makeatletter
\DeclareRobustCommand\onedot{\futurelet\@let@token\@onedot}
\def\@onedot{\ifx\@let@token.\else.\null\fi\xspace}

\def\eg{\emph{e.g}\onedot} 
\def\ie{\emph{i.e}\onedot}

\def\etal{\emph{et~al}\onedot}
\makeatother

\begin{document}

\title{Reading Text in the Wild with Convolutional Neural Networks}

\author{Max Jaderberg \and Karen Simonyan \and Andrea Vedaldi \and Andrew Zisserman }

\institute{Max Jaderberg \and Karen Simonyan \and Andrea Vedaldi \and Andrew Zisserman \at
	Department of Engineering Science, University of Oxford \\
	\email{\{max,karen,vedaldi,az\}@robots.ox.ac.uk}
}

\date{}

\maketitle

\begin{abstract}
In this work we present an end-to-end system for text spotting -- localising and recognising text in natural scene images -- and text based image retrieval. This system is based on a region proposal mechanism for detection and deep convolutional neural networks for recognition. Our pipeline uses a novel combination of complementary proposal generation techniques to ensure high recall, and a fast subsequent filtering stage for improving precision. For the recognition and ranking of proposals, we train very large convolutional neural networks to perform word recognition on the whole proposal region at the same time, departing from the character classifier based systems of the past. These networks are trained solely on data produced by a synthetic text generation engine, requiring no human labelled data. 

Analysing the stages of our pipeline, we show state-of-the-art performance throughout. We perform rigorous experiments across a number of standard end-to-end text spotting benchmarks and text-based image retrieval datasets, showing a large improvement over all previous methods. Finally, we demonstrate a real-world application of our text spotting system to allow thousands of hours of news footage to be instantly searchable via a text query.
\end{abstract}

\section{Introduction}
\label{sec:intro}

\begin{figure*}[t]
\centering
\includegraphics[width=\linewidth]{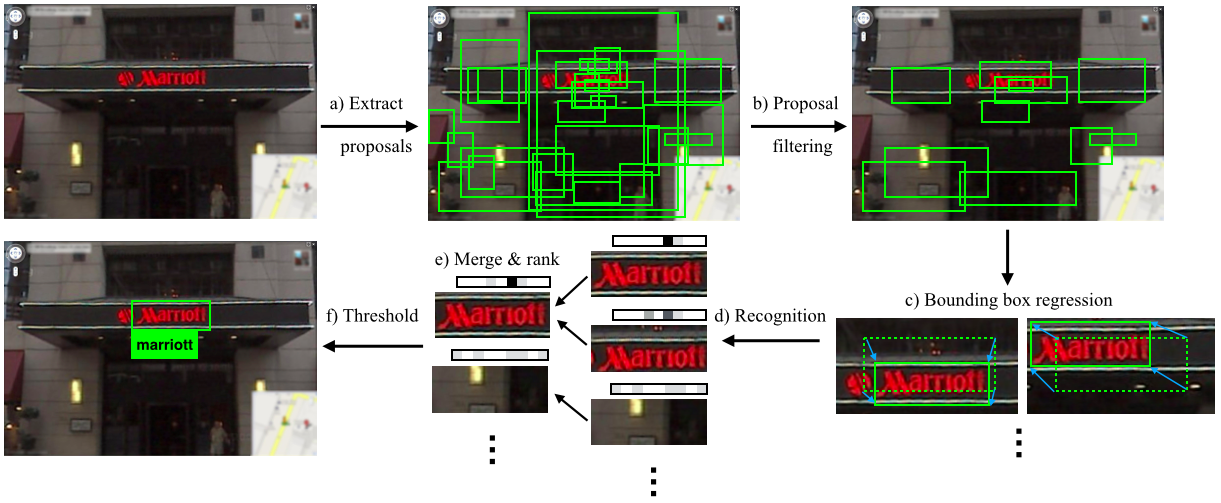}
\caption{The end-to-end text spotting pipeline proposed. a) A combination of region proposal methods extracts many word bounding box proposals. b) Proposals are filtered with a random forest classifier reducing number of false-positive detections. c) A CNN is used to perform bounding box regression for refining the proposals. d) A CNN performs text recognition on each of the refined proposals. e) Detections are merged based on proximity and recognition results and assigned a score. f) Thresholding the detections results in the final text spotting result.}
\label{fig:pipeline}
\end{figure*}

The automatic detection and recognition of text in natural images, \emph{text spotting}, is an important challenge for visual understanding. 

Text, as the physical incarnation of language, is one of the basic tools for preserving and communicating information. Much of the modern world is designed to be interpreted through the use of labels and other textual cues, and so text finds itself scattered throughout many images and videos. Through the use of text spotting, an important part of the semantic content of visual media can be decoded and used, for example, for understanding, annotating, and retrieving the billions of consumer photos produced every day.

Traditionally, text recognition has been focussed on document images, where OCR techniques are well suited to digitise planar, paper-based documents. However, when applied to natural scene images, these document OCR techniques fail as they are tuned to the largely black-and-white, line-based environment of printed documents. The text that occurs in natural scene images is hugely variable in appearance and layout, being drawn from a large number of fonts and styles, suffering from inconsistent lighting, occlusions, orientations, noise, and, in addition, the presence of background objects causes spurious false-positive detections. This places text spotting as a separate, far more challenging problem than document OCR.

The increase of powerful computer vision techniques and the overwhelming increase in the volume of images produced over the last decade has seen a rapid development of text spotting methods. To efficiently perform text spotting, the majority of methods follow the intuitive process of splitting the task in two: text detection followed by word recognition~\cite{Chen04text}. Text detection involves generating candidate character or word region detections, while word recognition takes these proposals and infers the words depicted.

In this paper we advance text spotting methods, making a number of key contributions as part of this.

Our main contribution is a novel text recognition method -- this is in the form of a deep convolutional neural network (CNN) \cite{Lecun98} which takes the \emph{whole word image as input} to the network. Evidence is gradually pooled from across the image to perform classification of the word across a huge dictionary, such as the 90k-word dictionary evaluated in this paper. Remarkably, our model is trained \emph{purely on synthetic data}, without incurring the cost of human labelling. We also propose an \emph{incremental learning} method to successfully train a model with such a large number of classes. Our recognition framework is exceptionally powerful, substantially outperforming previous state of the art on real-world scene text recognition, without using any real-world labelled training data.  

Our second contribution is a novel detection strategy for text spotting: the use of fast region proposal methods to perform word detection. We use a combination of an object-agnostic region proposal method and a sliding window detector. This gives very high recall coverage of individual word bounding boxes, resulting in around 98\% word recall on both ICDAR 2003 and Street View Text datasets with a manageable number of proposals. False-negative candidate word bounding boxes are filtered with a stronger random forest classifier and the remaining proposals adjusted using a CNN trained to regress the bounding box coordinates.

Our third contribution is the application of our pipeline for large-scale visual search of text in video. In a fraction of a second we are able to retrieve images and videos from a huge corpus that contain the visual rendering of a user given text query, at very high precision.

We expose the performance of each part of the pipeline in experiments, showing that we can maintain the high recall of the initial proposal stage while gradually boosting precision as more complex models and higher order information is incorporated. The recall of the detection stage is shown to be significantly higher than that of previous text detection methods, and the accuracy of the word recognition stage higher than all previous methods. The result is an end-to-end text spotting system that outperforms all previous methods by a large margin. We demonstrate this for the annotation task (localising and recognising text in images) across a large range of standard text spotting datasets, as well as in a retrieval scenario (retrieving a ranked list of images that contain the text of a query string) for standard datasets. In addition, the use of our framework for retrieval is further demonstrated in a real-world application -- being used to instantly search through thousands of hours of archived news footage for a user-given text query.

The following section gives an overview of our pipeline. We then review a selection of related work in Section~\ref{sec:related}. Sections~4-7 present the stages of our pipeline. We extensively test all elements of our pipeline in Section~\ref{sec:experiments} and include the details of datasets and the experimental setup. Finally, Section~\ref{sec:conclusions} summarises and concludes.

Our word recognition framework appeared previously as a tech report~\cite{Jaderberg14c} and at the NIPS 2014 Deep Learning and Representation Learning Workshop, along with some other non-dictionary based variants.

\section{Overview of the Approach}
\label{sec:approach}

The stages of our approach are as follows: word bounding box proposal generation (Section~\ref{sec:proposals}), proposal filtering and adjustments (Section~\ref{sec:filtering}), text recognition (Section~\ref{sec:recognition}) and final merging for the specific task (Section~\ref{sec:merging}). The full process is illustrated in Fig.~\ref{fig:pipeline}.

Our process loosely follows the detection/recognition separation -- a word detection stage followed by a word recognition stage. However, these two stages are not wholly distinct, as we use the information gained from word recognition to merge and rank detection results at the end, leading to a stronger holistic text spotting system.

The detection stage of our pipeline is based on weak-but-fast detection methods to generate word bounding-box proposals. This draws on the success of the R-CNN object detection framework of Girshick \etal \cite{Girshick14} where region proposals are mapped to a fixed size for CNN recognition. The use of region proposals avoids the computational complexity of evaluating an expensive classifier with exhaustive multi-scale, multi-aspect-ratio sliding window search. We use a combination of Edge Box proposals~\cite{Zitnick14} and a trained aggregate channel features detector~\cite{Dollar14} to generate candidate word bounding boxes. Due to the large number of false-positive proposals, we then use a random forest classifier to filter the number of proposals to a manageable size -- this is a stronger classifier than those found in the proposal algorithms. Finally, inspired by the success of bounding box regression in DPM \cite{Felzenszwalb10a} and R-CNN \cite{Girshick14}, we regress more accurate bounding boxes from the seeds of the proposal algorithms which greatly improves the average overlap ratio of positive detections with groundtruth. However, unlike the linear regressors of \cite{Felzenszwalb10a,Girshick14} we train a CNN specifically for regression. We discuss these design choices in each section.

The second stage of our framework produces a text recognition result for each proposal generated from the detection stage. We take a whole-word approach to recognition, providing the entire cropped region of the word as input to a deep convolutional neural network. We present a dictionary model which poses the recognition task as a multi-way classification task across a dictionary of 90k possible words. Due to the mammoth training data requirements of classification tasks of this scale, these models are trained \emph{purely from synthetic data}. Our synthetic data engine is capable of rendering sufficiently realistic and variable word image samples that the models trained on this data translate to the domain of real-world word images giving state-of-the-art recognition accuracy. 

Finally, we use the information gleaned from recognition to update our detection results with multiple rounds of non-maximal suppression and bounding box regression.

\section{Related Work}
\label{sec:related}
In this section we review the contributions of works most related to ours. These focus solely on text detection \cite{Anthimopoulos11, Chen11a,Epshtein10,Yi11,Yin13,Huang14}, text recognition \cite{Mishra12,Novikova12,Rath07,Bissacco13,Jaderberg14c,Almazan14,Yao14}, or on combining both in end-to-end
systems \cite{QuackText09,Posner10,Wang12,NeumannText10, Neumann11,Neumann12,Wang11,Neumann13,Alsharif13,Weinman13,Jaderberg14a,Gordo14,Mishra13}.

\subsection{Text Detection Methods}
Text detection methods tackle the first task of the standard text spotting pipeline~\cite{Chen04text}: producing segmentations or bounding boxes of words in natural scene images. Detecting instances of words in noisy and cluttered images is a highly non-trivial task, and the methods developed to solve this are based on either character regions~\cite{Chen11a,Epshtein10,Yi11,Yin13,NeumannText10, Neumann11,Neumann12,Neumann13,Huang14} or sliding windows~\cite{Jaderberg14a,QuackText09,Anthimopoulos11,Posner10,Wang11,Wang12}.

\emph{Character region methods} aim to segment pixels into characters, and then group characters into words. Epshtein \etal~\cite{Epshtein10} find regions of the input image which have constant stroke width -- the distance between two parallel edges -- by taking the stroke width transform (SWT). Intuitively characters are regions of similar stroke width, so clustering pixels together forms characters, and characters are grouped together into words based on geometric heuristics. In \cite{Neumann13}, Neumann and Matas revisit the notion of characters represented as strokes and use gradient filters to detect oriented strokes in place of the SWT. Rather than regions of constant stroke width, Neumann and Matas~\citep{Neumann11,NeumannText10,Neumann12} use Extremal Regions~\cite{Matas02} as character regions. Huang~\etal \cite{Huang14} expand on the use of Maximally Stable Extremal Regions by incorporating a strong CNN classifier to efficiently prune the trees of Extremal Regions leading to less false-positive detections. 

\emph{Sliding window methods} approach text detection as a classical object detection task. Wang~\etal~\cite{Wang11} use a random ferns~\cite{Ozuysal07} classifier trained on HOG features~\cite{Felzenszwalb10a} in a sliding window scenario to find characters in an image. These are grouped into words using a pictorial structures framework~\cite{Felzenszwalb05} for a small fixed lexicon. Wang~\&~Wu \etal \cite{Wang12} show that CNNs trained for character classification can be used as effective sliding window classifiers. In some of our earlier work~\cite{Jaderberg14a}, we use CNNs for text detection by training a text/no-text classifier for sliding window evaluations, and also CNNs for character and bigram classification to perform word recognition. We showed that using feature sharing across all the CNNs for the different classification tasks resulted in stronger classifiers for text detection than training each classifier independently.

Unlike previous methods, our framework operates in a low-precision, high-recall mode -- rather than using a single word location proposal, we carry a sufficiently high number of candidates through several stages of our pipeline. We use high recall region proposal methods and a filtering stage to further refine these. In fact, our ``detection method'' is only complete after performing full text recognition on each remaining proposal, as we then merge and rank the proposals based on the output of the recognition stage to give the final detections, complete with their recognition results.

\subsection{Text Recognition Methods}
Text recognition aims at taking a cropped image of a single word and recognising the word depicted. While there are many previous works focussing on handwriting or historical document recognition~\cite{Fischer10,Rath07,Frinken12,Manmatha96}, these methods don't generalise in function to generic scene text due to the highly variable foreground and background textures that are not present with documents. 

For scene text recognition, methods can be split into two groups -- character based recognition~\cite{Yao14,Jaderberg14a,Bissacco13,Alsharif13,Wang11,Wang12,QuackText09,Posner10,Weinman13} and whole word based recognition~\cite{Goel13,Rodriguez13,Novikova12,Mishra12,Almazan14,Jaderberg14c}. 

\emph{Character based recognition} relies on an individual character classifier for per-character recognition which is integrated across the word image to generate the full word recognition. In \cite{Yao14}, Yao \etal learn a set of mid-level features, strokelets, by clustering sub-patches of characters. Characters are detected with Hough voting, with the characters identified by a random forest classifier acting on strokelet and HOG features. 

The works of \cite{Wang12,Jaderberg14a,Bissacco13,Alsharif13} all use CNNs as character classifiers. \cite{Bissacco13} and \cite{Alsharif13} over-segment the word image into potential character regions, either through unsupervised binarization techniques or with a supervised classifier. Alsharif \etal \cite{Alsharif13} then use a complicated combination of segmentation-correction and character recognition CNNs together with an HMM with a fixed lexicon to generate the final recognition result. The PhotoOCR system \cite{Bissacco13} uses a neural network classifier acting on the HOG features of the segments as scores to find the best combination of segments using beam search. The beam search incorporates a strong N-gram language model, and the final beam search proposals are re-ranked with a further language model and shape model. Our own previous work~\cite{Jaderberg14a} uses a combination of a binary text/no-text classifier, a character classifier, and a bigram classifier densely computed across the word image as cues to a Viterbi scoring function in the context of a fixed lexicon.

As an alternative approach to word recognition other methods use \emph{whole word based recognition}, pooling features from across the entire word sub-image before performing word classification. The works of Mishra \etal~\cite{Mishra12} and Novikova \etal~\cite{Novikova12} still
rely on explicit character classifiers, but construct a graph to infer
the word, pooling together the full word
evidence. Goel \etal~\cite{Goel13} use whole word sub-image features to recognise
words by comparing to simple black-and-white font-renderings of
lexicon words. Rodriguez~\etal~\cite{Rodriguez13} use aggregated Fisher
Vectors~\cite{Perronnin10} and a Structured SVM
framework to create a joint word-image and text
embedding.

Almazan \etal \cite{Almazan14} further explore the notion of word embeddings, creating a joint embedding space for word images and representations of word strings. This is extended in \cite{Gordo14} where Gordo makes explicit use of character level training data to learn mid-level features. This results in performance on par with \cite{Bissacco13} but using only a small fraction of the amount of training data. 

While not performing full scene text recognition, Goodfellow~\etal~\cite{Goodfellow13} had great success using a CNN with multiple position-sensitive character classifier outputs to perform street number recognition. This model was extended to CAPTCHA sequences up to 8 characters long where they demonstrated impressive performance using synthetic training data for a synthetic problem (where the generative model is known). In contrast, we show that synthetic training data can be used for a real-world data problem (where the generative model is unknown).

Our method for text recognition also follows a whole word image approach. Similarly to \cite{Goodfellow13}, we take the word image as input to a deep CNN, however we employ a dictionary classification model. Recognition is achieved by performing multi-way classification across the entire dictionary of potential words.

In the following sections we describe the details of each stage of our text spotting pipeline. The sections are presented in order of their use in the end-to-end system.

\section{Proposal Generation}
\label{sec:proposals}

The first stage of our end-to-end text spotting pipeline relies on the generation of word bounding boxes. This is word detection -- in an ideal scenario we would be able to generate word bounding boxes with high recall and high precision, achieving this by extracting the maximum amount of information from each bounding box candidate possible. However, in practice a precision/recall tradeoff is required to reduce computational complexity. With this in mind we opt for a fast, high recall initial phase, using computationally cheap classifiers, and gradually incorporate more information and more complex models to improve precision by rejecting false-positive detections resulting in a cascade. To compute recall and precision in a detection scenario, a bounding box is said to be a true-positive detection if it has overlap with a groundtruth bounding box above a defined threshold. The overlap for bounding boxes $b_1$ and $b_2$ is defined as the ratio of intersection over union (IoU): $\frac{|b_1 \cap b_2|}{|b_1 \cup b_2|}$.

Though never applied to word detection before, region proposal methods have gained a lot of attention for generic object detection. Region proposal methods~\cite{Uijlings13,Alexe12, Cheng14,Zitnick14} aim to generate object region proposals with high recall, but at the cost of a large number of false-positive detections. Even so, this still reduces the search space drastically compared to sliding window evaluation of the subsequent stages of a detection pipeline. Effectively, region proposal methods can be viewed as a weak detector.

In this work we combine the results of two detection mechanisms -- the Edge Boxes region proposal algorithm (\cite{Zitnick14}, Section~\ref{sec:edgeboxes}) and a weak aggregate channel features detector (\cite{Dollar14}, Section~\ref{sec:detector}). 

\subsection{Edge Boxes}
\label{sec:edgeboxes}
We use the formulation of Edge Boxes as described in~\cite{Zitnick14}. The key intuition behind Edge Boxes is that since objects are generally self contained, the number of contours wholly enclosed by a bounding box is indicative of the likelihood of the box containing an object. Edges tend to correspond to object boundaries, and so if edges are contained inside a bounding box this implies objects are contained within the bounding box, whereas edges which cross the border of the bounding box suggest there is an object that is not wholly contained by the bounding box.

The notion of an object being a collection of boundaries is especially true when the desired objects are words -- collections of characters with sharp boundaries.

Following~\cite{Zitnick14}, we compute the edge response map using the Structured Edge detector~\cite{Dollar13,Dollar14b} and perform Non-Maximal Suppression orthogonal to the edge responses, sparsifying the edge map. A candidate bounding box $b$ is assigned a score $s_b$ based on the number of edges wholly contained by $b$, normalised by the perimeter of $b$. The full details can be found in~\cite{Zitnick14}.

The boxes $b$ are evaluated in a sliding window manner, over multiple scales and aspect ratios, and given a score $s_b$. Finally, the boxes are sorted by score and non-maximal suppression is performed: a box is removed if its overlap with another box of higher score is more than a threshold. This results in a set of candidate bounding boxes for words $B_e$.

\subsection{Aggregate Channel Feature Detector}
\label{sec:detector}
Another method for generating candidate word bounding box proposals is by using a conventional trained detector. We use the aggregate channel features (ACF) detector framework of~\cite{Dollar14} for its speed of computation. This is a conventional sliding window detector based on ACF features coupled with an AdaBoost classifier. ACF based detectors have been shown to work well on pedestrian detection and general object detection, and here we use the same framework for word detection.

For each image $I$ a number of feature channels are computed, such that channel $C = \Omega(I)$, where $\Omega$ is the channel feature extraction function. We use channels similar to those in~\cite{Dollar10}: normalised gradient magnitude, histogram of oriented gradients (6 channels), and the raw greyscale input. Each channel $C$ is smoothed, divided into blocks and the pixels in each block are summed and smoothed again, resulting in aggregate channel features. 

The ACF features are not scale invariant, so for multi-scale detection we need to extract features at many different scales -- a feature pyramid. In a standard detection pipeline, the channel features for a particular scale $s$ are computed by resampling the image and recomputing the channel features $C_s = \Omega(I_s)$ where $C_s$ are the channel features at scale $s$ and $I_s = R(I,s)$ is the image resampled by $s$. Resampling and recomputing the features at every scale is computationally expensive. However, as shown in~\cite{Dollar10,Dollar14}, the channel features at scale $s$ can be approximated by resampling the features at a different scale, such that $C_s \approx R(C,s) \cdot s^{-\lambda_\Omega}$, where $\lambda_\Omega$ is a channel specific power-law factor. Therefore, fast feature pyramids can be computed by evaluating $C_s = \Omega(R(I,s))$ at only a single scale per octave ($s \in \{1, \frac{1}{2}, \frac{1}{4}, \dots\}$) and at intermediate scales, $C_s$ is computed using $C_s = R(C_{s'},s/s')(s/s')^{-\lambda_\Omega}$ where $s' \in \{1, \frac{1}{2}, \frac{1}{4}, \dots\}$. This results in much faster feature pyramid computation.

The sliding window classifier is an ensemble of weak decision trees, trained using AdaBoost~\cite{Friedman00}, using the aggregate channel features. We evaluate the classifier on every block of aggregate channel features in our feature pyramid, and repeat this for multiple aspect ratios to account for different length words, giving a score for each box. Thresholding on score gives a set of word proposal bounding boxes from the detector, $B_d$.

\paragraph{Discussion.}
We experimented with a number of region proposal algorithms. Some were too slow to be useful~\cite{Uijlings13,Alexe12}. A very fast method, BING~\cite{Cheng14}, gives good recall when re-trained specifically for word detection but achieves a low overlap ratio for detections and poorer overall recall. We found Edge Boxes to give the best recall and overlap ratio.

It was also observed that independently, neither Edge Boxes nor the ACF detector achieve particularly high recall, 92\% and 70\% recall respectively (see Section~\ref{sec:implementation} for experiments), but when the proposals are combined achieve 98\% recall (recall is computed at 0.5 overlap). In contrast, combining BING, with a recall of 86\% with the ACF detector gives a combined recall of only 92\%. This suggests that the Edge Box and ACF detector methods are very complementary when used in conjunction, and so we compose the final set of candidate bounding boxes $B = \{B_e \cup B_d\}$.

\section{Filtering \& Refinement}
\label{sec:filtering}
The proposal generation stage of Section~\ref{sec:proposals} produces a set of candidate bounding boxes $B$. However, to achieve a high recall, thousands of bounding boxes are generated, most of which are false-positive. We therefore aim to use a stronger classifier to further filter these to a number that is computationally manageable for the more expensive full text recognition stage described in Section~\ref{sec:classification}. We also observe that the overlap of many of the bounding boxes with the groundtruth is unsatisfactorily low, and therefore train a regressor to refine the location of the bounding boxes, as described in Section~\ref{sec:regression}. 

\subsection{Word Classification}
\label{sec:classification}
To reduce the number of false-positive word detections, we seek a classifier to perform word/no-word binary classification. For this we use a random forest classifier~\cite{Breiman01} acting on HOG features~\cite{Felzenszwalb10a}. 

For each bounding box proposal $b \in B$ we resample the cropped image region to a fixed size and extract HOG features, resulting in a descriptor $h$. The descriptor is then classified with a random forest classifier, with decision stump nodes. The random forest classifies every proposal, and the proposals falling below a certain threshold are rejected, leaving a filtered set of bounding boxes $B_f$.

\subsection{Bounding Box Regression}
\label{sec:regression}

\begin{figure}
\begin{center}
\includegraphics[width=0.7\linewidth]{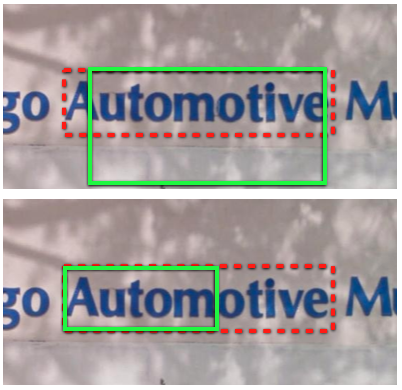} 
\caption{The shortcoming of using an overlap ratio for text detection of 0.5. The two examples of proposal bounding boxes (green solid box) have approximately 0.5 overlap with groundtruth (red dashed box). In the bottom case, a 0.5 overlap is not satisfactory to produce accurate text recognition results.}
\label{fig:overlap}
\end{center}
\end{figure}

Although our proposal mechanism and filtering stage give very high recall, the overlap of these proposals can be quite poor. 

While an overlap of 0.5 is usually acceptable for general object detection~\cite{Everingham10}, for accurate text recognition this can be unsatisfactory. This is especially true when one edge of the bounding box is predicted accurately but not the other -- \eg if the height of the bounding box is computed perfectly, the width of the bounding box can be either double or half as wide as it should be and still achieve 0.5 overlap. This is illustrated in Fig.~\ref{fig:overlap}. Both predicted bounding boxes have 0.5 overlap with the groundtruth, but for text this can amount to only seeing half the word if the height is correctly computed, and so it would be impossible to recognise the correct word in the bottom example of Fig.~\ref{fig:overlap}. Note that both proposals contain text, so neither are filtered by the word/no-word classifier.

Due to the large number of region proposals, we could hope that there would be some proposals that would overlap with the groundtruth to a satisfactory degree, but we can encourage this by explicitly refining the coordinates of the proposed bounding boxes -- we do this within a regression framework. 

\begin{figure}
\begin{center}
\includegraphics[width=1.0\linewidth]{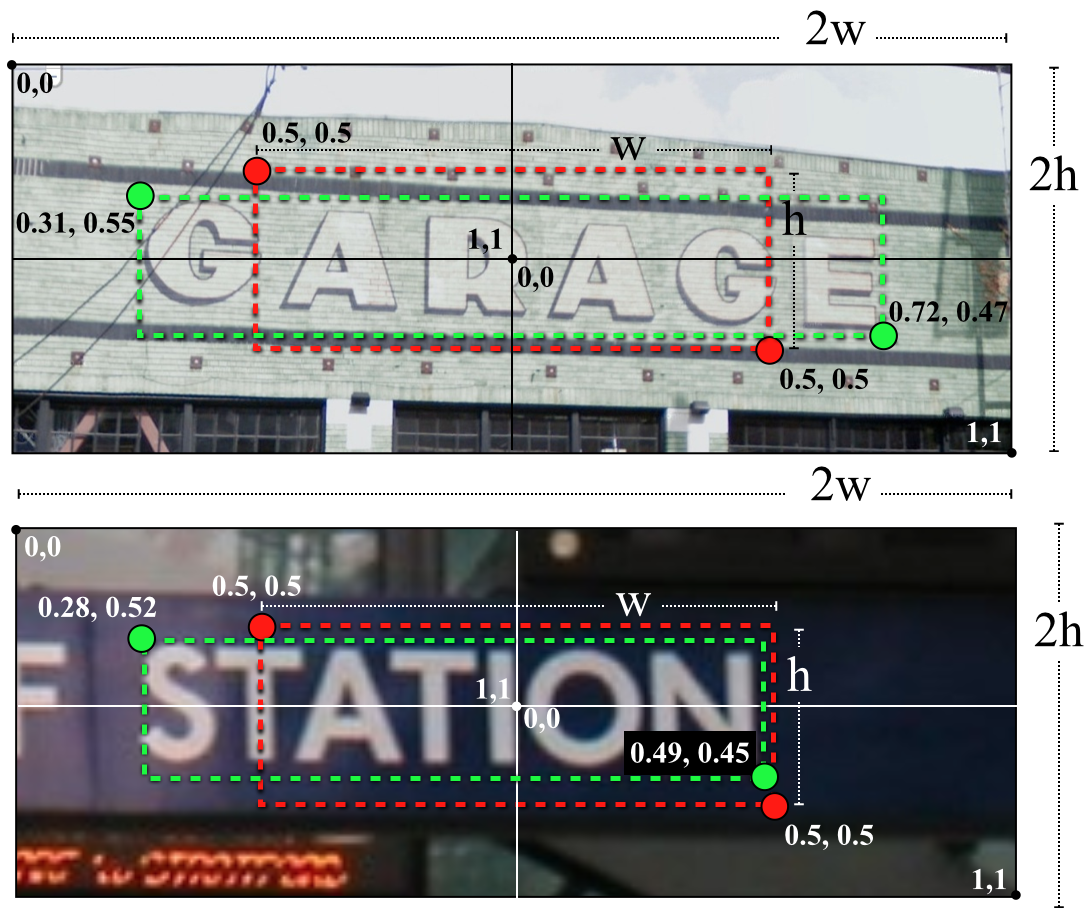} 
\caption{The bounding box regression encoding scheme showing the original proposal (red) and the adjusted proposal (green). The cropped input image shown is always centred on the original proposal, meaning the original proposal always has implied encoded coordinates of $(0.5,0.5,0.5,0.5)$.}
\label{fig:encoding}
\end{center}
\end{figure}

Our bounding box coordinate regressor takes each proposed bounding box $b \in B_f$ and produces an updated estimate of that proposal $b^*$. A bounding box is parametrised by its top-left and bottom-right corners, such that bounding box $b = (x_1,y_1,x_2,y_2)$. The full image $I$ is cropped to a rectangle centred on the region $b$, with the width and height inflated by a scale factor. The resulting image is resampled to a fixed size $W \times H$, giving $I_b$, which is processed by the CNN to regress the four values of $b^*$. We do not regress the absolute values of the bounding box coordinates directly, but rather encoded values. The top-left coordinate is encoded by the top-left quadrant of $I_b$, and the bottom left coordinate by the bottom-left quadrant of $I_b$ as illustrated by Fig.~\ref{fig:encoding}. This normalises the coordinates to generally fall in the interval $[0,1]$, but allows the breaking of this interval if required.

In practice, we inflate the cropping region of each proposal by a factor of two. This gives the CNN enough context to predict a more accurate location of the proposal bounding box. The CNN is trained with example pairs of $(I_b,b_{gt})$ to regress the groundtruth bounding box $b_{gt}$ from the sub-image $I_b$ cropped from $I$ by the estimated bounding box $b$. This is done by minimising the $L_2$ loss between the encoded bounding boxes, \ie 
\begin{equation}
\min_{\Phi} \sum_{b\in B_{train}} \|g(I_b;\Phi) - q(b_{gt})\|^2_2
\label{eqn:regcost}
\end{equation}
over the network parameters $\Phi$ on a training set $B_{train}$, where $g$ is the CNN forward pass function and $q$ is the bounding box coordinate encoder.

\paragraph{Discussion.}
The choice of which features and the classifier and regression methods to use was made through experimenting with a number of different choices. This included using a CNN for classification, with a dedicated classification CNN, and also by jointly training a single CNN to perform both classification and regression simultaneously with multi-task learning. However, the classification performance of the CNN was not significantly better than that of HOG with a random forest, but requires more computations and a GPU for processing. We therefore chose the random forest classifier to reduce the computational cost of our pipeline without impacting end results.

The bounding box regression not only improves the overlap to aid text recognition for each individual sample, but also causes many proposals to converge on the same bounding box coordinates for a single instance of a word, therefore aiding the voting/merging mechanism described in Section~\ref{sec:textspotting} with duplicate detections. 

\section{Text Recognition}
\label{sec:recognition}

At this stage of our processing pipeline, a pool of accurate word bounding box proposals has been generated as described in the previous sections. We now turn to the task of recognising words inside these proposed bounding boxes. To this end we use a deep CNN to perform classification across a pre-defined dictionary of words -- dictionary encoding -- which explicitly models natural language. The cropped image of each of the proposed bounding boxes is taken as input to the CNN, and the CNN produces a probability distribution over all the words in the dictionary. The word with the maximum probability can be taken as the recognition result. 

The model, described fully in Section~\ref{sec:model}, can scale to a huge dictionary of 90k words, encompassing the majority of the commonly used English language (see Section~\ref{sec:datasets} for details of the dictionary used). However, to achieve this, many training samples of every different possible word must be amassed. Such a training dataset does not exist, so we instead use synthetic training data, described in Section~\ref{sec:synthdata}, to train our CNN. This synthetic data is so realistic that the CNN can be trained purely on the synthetic data but still applied to real world data.

\subsection{Synthetic Training Data}
\label{sec:synthdata}

\begin{figure*}
\centering
\begin{tabular}{>{\centering}m{10pt}m{0.95\textwidth}}
(a) & \includegraphics[width=0.95\textwidth]{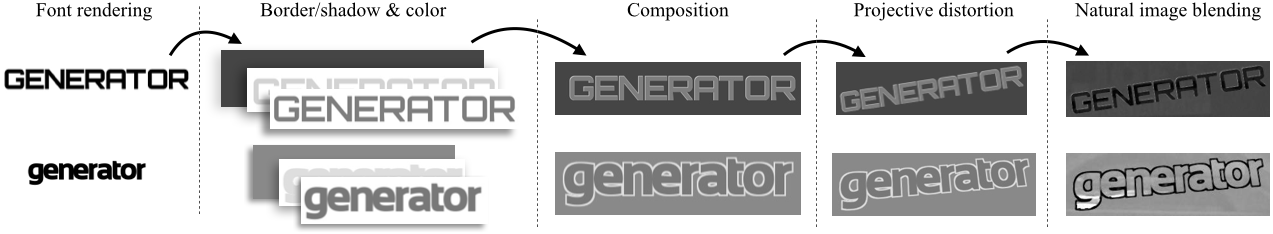}\\
\hline
(b) & \includegraphics[width=0.95\textwidth,trim=0 0 0 -10pt]{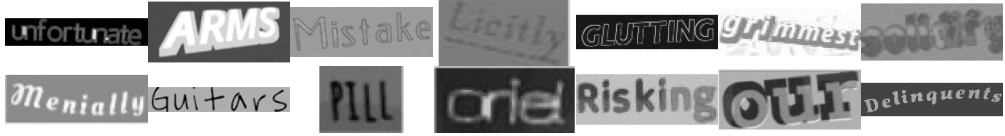}\\
\end{tabular}
\vspace*{-3mm}
\caption{(a) The text generation process after font rendering, creating and colouring the image-layers, applying projective distortions, and after image blending. (b) Some randomly sampled data created by the synthetic text engine.}
\label{fig:synthdata}
\end{figure*}

This section describes our scene text rendering algorithm. As our CNN models take whole word images as input instead of individual character images, it is essential to have access to a training dataset of cropped word images that covers the whole language or at least a target lexicon. While there are some publicly available datasets from ICDAR~\cite{ICDAR03, ICDAR2005, ICDAR11, ICDAR2013}, the Street View Text (SVT) dataset~\cite{Wang11}, the IIIT-5k dataset~\cite{Mishra12}, and others, the number of full word image samples is only in the thousands, and the vocabulary is very limited. 

The lack of full word image samples has caused previous work to rely on character classifiers instead (as character data is plentiful), or this deficit in training data has been mitigated by mining for data or having access to large proprietary datasets~\cite{Jaderberg14a,Bissacco13,Goodfellow13}. However, we wish to perform whole word image based recognition and move away from character recognition, and aim to do this in a scalable manner without requiring human labelled datasets.

Following the success of some synthetic character datasets~\cite{Campos09,Wang12}, we create a synthetic word data generator, capable of emulating the distribution of scene text images. This is a reasonable goal, considering that much of the text found in natural scenes is restricted to a limited set of computer-generated fonts, and only the physical rendering process (\eg printing, painting) and the imaging process (\eg camera, viewpoint, illumination, clutter) are not controlled by a computer algorithm. 

Fig.~\ref{fig:synthdata} illustrates the generative process and some resulting synthetic data samples. These samples are composed of three separate image-layers -- a background image-layer, foreground image-layer, and optional border/shadow image-layer -- which are in the form of an image with an alpha channel. The synthetic data generation process is as follows:
\begin{enumerate}
  \item {\emph{Font rendering} -- a font is randomly selected from a catalogue of over 1400 fonts downloaded from Google Fonts. The kerning, weight, underline, and other properties are varied randomly from arbitrarily defined distributions. The word is rendered on to the foreground image-layer's alpha channel with either a horizontal bottom text line or following a random curve.}
  \item \emph{Border/shadow rendering} -- an inset border, outset border, or shadow with a random width may be rendered from the foreground.

\item \emph{Base colouring} -- each of the three image-layers are 
filled with a different uniform colour sampled
from clusters over natural images. The clusters are formed by
k-means clustering the RGB components of each image of the
training datasets of~\cite{ICDAR03} into three clusters.

  \item \emph{Projective distortion} -- the foreground and border/shadow image-layers are distorted with a random, full projective transformation, simulating the 3D world.
  \item \emph{Natural data blending} -- each of the image-layers are blended with a randomly-sampled crop of an image from the training datasets of ICDAR 2003 and SVT. The amount of blend and alpha blend mode (\eg~normal, add, multiply, burn, max, \emph{etc.}) is dictated by a random process, and this creates an eclectic range of textures and compositions. The three image-layers are also blended together in a random manner, to give a single output image.
  \item \emph{Noise} -- Elastic distortion similar to \cite{Simard03}, Gaussian noise, blur, resampling noise, and JPEG compression artefacts are introduced to the image.
\end{enumerate}

This process produces a wide range of synthetic data samples, being drawn from a multitude of random distributions, mimicking real-world samples of scene text images. The synthetic data is used in place of real-world data, and the labels are generated from a corpus or dictionary as desired. By creating training datasets many orders of magnitude larger than what has been available before, we are able to use data-hungry deep learning algorithms to train a richer, whole-word-based model.

\subsection{CNN Model}
\label{sec:model}

\begin{figure*}[t]
\centering
\includegraphics[width=0.9\linewidth]{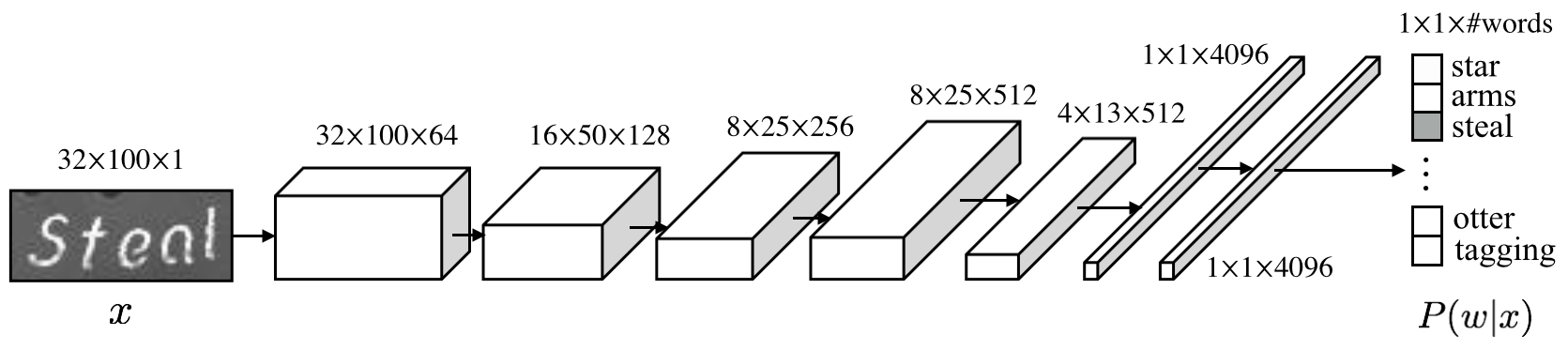}
\caption{A schematic of the CNN used for text recognition by word classification. The dimensions of the featuremaps at each layer of the network are shown.}
\label{fig:dictnet}
\end{figure*}

\newcommand{\dict}{\mathcal{W}}
This section describes our model for word recognition. We formulate recognition as a multi-class classification problem, with one class per word, where words $w$ are constrained to be selected in a pre-defined dictionary $\dict$. While the dictionary $\dict$ of a natural language may seem too large for this approach to be feasible, in practice an advanced English vocabulary, including different word forms, contains only around 90k words, which is large but manageable.

In detail, we propose to use a CNN classifier where each word $w\in\dict$ in the lexicon corresponds to an output neuron. We use a CNN with five convolutional layers and three fully-connected layers, with the exact details described in Section~\ref{sec:implementation}. The final fully-connected layer performs classification across the dictionary of words, so has the same number of units as the size of the dictionary we wish to recognise. 

The predicted word recognition result $w^*$ out of the set of all dictionary words $\dict$ in a language $\mathcal{L}$ for a given input image $x$ is given by 
\begin{equation}
w^* = \arg\max_{w \in \dict} P(w|x,\mathcal{L}). 
\label{eqn:word}
\end{equation}
Since $P(w|x,\mathcal{L})$ can be written as
\begin{equation}
P(w|x,\mathcal{L}) = \frac{P(w|x)P(w|\mathcal{L})P(x)}{P(x|\mathcal{L})P(w)}
\end{equation}
and with the assumptions that $x$ is independent of $\mathcal{L}$ and that prior to any knowledge of our language all words are equally probable, our scoring function reduces to 
\begin{equation}
w^* = \arg\max_{w \in \dict} P(w|x)P(w|\mathcal{L}).
\end{equation}
The per-word output probability $P(w|x)$ is modelled by the softmax output of the final fully-connected layer of the recognition CNN, and the language based word prior $P(w|\mathcal{L})$ can be modelled by a lexicon or frequency counts. A schematic of the network is shown in Fig.~\ref{fig:dictnet}.

One limitation of this CNN model is that the input $x$ must be a fixed, pre-defined size. This is problematic for word images, as although the height of the image is always one character tall, the width of a word image is highly dependent on the number of characters in the word, which can range between one and 23 characters. To overcome this issue, we simply resample the word image to a fixed width and height. Although this does not preserve the aspect ratio, the horizontal frequency distortion of image features most likely provides the network with word-length cues. We also experimented with different padding regimes to preserve the aspect ratio, but found that the results are not quite as good as performing naive resampling. 

To summarise, for each proposal bounding box $b \in B_f$ for image $I$ we compute $P(w|x_b,\mathcal{L})$ by cropping the image to $I_b = c(b,I)$, resampling to fixed dimensions $W \times H$ such that $x_b = R(I_b, W, H)$, and compute $P(w|x_b)$ with the text recognition CNN and multiply by $P(w|\mathcal{L})$ (task dependent) to give a final probability distribution over words $P(w|x_b,\mathcal{L})$.

\section{Merging \& Ranking}
\label{sec:merging}
At this point in the pipeline, we have a set of word bounding boxes for each image $B_f$ with their associated word probability distributions $P_{B_f} = \{p_b : b \in B_f\}$, where $p_b = P(w|b,I) =  P(w|x_b,\mathcal{L})$. However, this set of detections still contains a number of false-positive and duplicate detections of words, so a final merging and ranking of detections must be performed depending on the task at hand: text spotting or text based image retrieval.

\subsection{Text Spotting}
The goal of text spotting is to localise and recognise the individual words in the image. Each word should be labelled by a bounding box enclosing the word and the bounding box should have an associated text label.

For this task, we assign each bounding box in $b \in B_f$ a label $w_b$ and score $s_b$ according to $b$'s maximum word probability:
\begin{equation}
w_b = \arg\max_{w \in \dict} P(w|b,I),~~s_b = \max_{w \in \dict} P(w|b,I)
\end{equation}

To cluster duplicate detections of the same word instance, we perform a greedy non maximum suppression (NMS) on detections with \emph{the same word} label, aggregating the scores of suppressed proposals. This can be seen as positional voting for a particular word. Subsequently, we perform NMS to suppress non-maximal detections of \emph{different words} with some overlap.

Our text recognition CNN is able to accurately recognise text in very loosely cropped word sub-images. Because of this, we find that some valid text spotting results have less than 0.5 overlap with groundtruth, but we require greater than 0.5 overlap for some applications (see Section~\ref{sec:textspotting}). 

\begin{figure*}
\begin{center}
\includegraphics[width=0.85\linewidth]{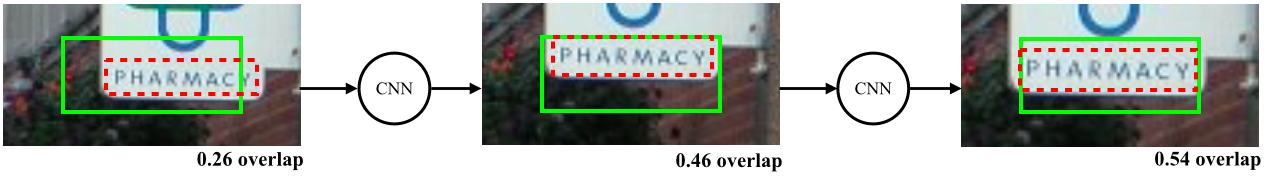} 
\caption{An example of the improvement in localisation of the word detection \texttt{pharmacy} through multiple rounds of recurrent regression.}
\label{fig:recreg}
\end{center}
\end{figure*}

To improve the overlap of detection results, we additionally perform multiple rounds of bounding box regression as in Section~\ref{sec:regression} and NMS as described above to further refine our detections. This can be seen as a recurrent regressor network. Each round of regression updates the prediction of the each word's localisation, giving the next round of regression an updated context window to perform the next regression, as shown in Fig.~\ref{fig:recreg}. Performing NMS between each regression causes bounding boxes that have become similar after the latest round of regression to be grouped as a single detection. This generally causes the overlap of detections to converge on a higher, stable value with only a few rounds of recurrent regression.

The refined results, given by the tuple $(b, w_b, s_b)$, are ranked by their scores $s_b$ and a threshold determines the final text spotting result. For the direct comparison of scores across images, we normalise the scores of the results of each image by the maximum score for a detection in that image.

\subsection{Image Retrieval}
For the task of text based image retrieval, we wish to retrieve the list of images which contain the given query words. Localisation of the query word is not required, only optional for giving evidence for retrieving that image. 

This is achieved by, at query time, assigning each image $I$ a score $s^\mathcal{Q}_I$ for the query words $\mathcal{Q} = \{q_1,q_2,\ldots\}$, and sorting the images in the database $\mathcal{I}$ in descending order of score. It is also required that the score for all images can be computed fast enough to scale to databases of millions of images, allowing fast retrieval of visual content by text search. While retrieval is often performed for just a single query word ($\mathcal{Q} = \{q\}$), we generalise our retrieval framework to be able to handle multiple query words.

We estimate the per-image probability distribution across word space $P(w|I)$ by averaging the word probability distributions across all detections $B_f$ in an image
\begin{equation}
p_I = P(w|I) = \frac{1}{|B_f|}\sum_{b \in B_f} p_b.
\end{equation}
This distribution is computed offline for all $I \in \mathcal{I}$.

At query time, we can simply compute a score for each image $s^\mathcal{Q}_I$ representing the probability that the image $I$ contains any of the query words $\mathcal{Q}$. Assuming independence between the presence of query words
\begin{equation}
s^\mathcal{Q}_I = \sum_{q \in \mathcal{Q}} P(q|I) = \sum_{q \in \mathcal{Q}} p_I(q)
\end{equation}
where $p_I(q)$ is just a lookup of the probability of word $q$ in the word distribution $p_I$. 
These scores can be computed very quickly and efficiently by constructing an inverted index of $p_I~\forall~I \in \mathcal{I}$.

After a one-time, offline pre-processing to compute $p_I$ and assemble the inverted index, a query can be processed across a database of millions of images in less than a second.

\section{Experiments}
\label{sec:experiments}

In this section we evaluate our pipeline on a number of standard text spotting and text based image retrieval benchmarks. 

We introduce the various datasets used for evaluation in Section~\ref{sec:datasets}, give the exact implementation details and results of each part of our pipeline in Section~\ref{sec:implementation}, and finally present the results on text spotting and image retrieval benchmarks in Section~\ref{sec:textspotting} and Section~\ref{sec:retrieval} respectively.

\subsection{Datasets}
\label{sec:datasets}
We evaluate our pipeline on an extensive number of datasets. Due to different levels of annotation, the datasets are used for a combination of text recognition, text spotting, and image retrieval evaluation. The datasets are summarised in Table~\ref{table:evalrecog}, Table~\ref{table:evalspotting}, and Table~\ref{table:evalretrieval}. The smaller lexicons provided by some datasets are used to reduce the search space to just text contained within the lexicons.

{\bf The Synth dataset}
 is generated by our synthetic data engine of Section~\ref{sec:synthdata}. We generate 9 million $32 \times 100$ images, with equal numbers of word samples from a 90k word dictionary. We use 900k of these for a testing dataset, 900k for validation, and the remaining for training. The 90k dictionary consists of the English dictionary from Hunspell~\cite{hunspell}, a popular open source spell checking system. This dictionary consists of 50k root words, and we expand this to include all the prefixes and suffixes possible, as well as adding in the test dataset words from the ICDAR, SVT and IIIT datasets -- 90k words in total. \emph{This dataset is publicly available at } \url{http://www.robots.ox.ac.uk/~vgg/data/text/}.

{\bf ICDAR 2003 (IC03)~\cite{icdar2003dataset}, ICDAR 2011 (IC11) \cite{ICDAR11}, and ICDAR 2013 (IC13)~\cite{ICDAR2013}}
are scene text recognition datasets consisting of 251, 255, and 233 full scene images respectively. The photos consist of a range of scenes and word level annotation is provided. Much of the test data is the same between the three datasets. For IC03, Wang~\cite{Wang11} defines per-image 50 word lexicons (IC03-50) and a lexicon of all test groundtruth words (IC03-Full). For IC11, \cite{Mishra13} defines a list of 538 query words to evaluate text based image retrieval.

{\bf The Street View Text (SVT) dataset~\cite{Wang11}}
consists of 249 high resolution images downloaded from Google StreetView of road-side scenes. This is a challenging dataset with a lot of noise, as well as suffering from many unannotated words. Per-image 50 word lexicons (SVT-50) are also provided.

{\bf The IIIT 5k-word dataset~\cite{Mishra12}}
contains 3000 cropped word images of scene text and digital images obtained from Google image search. This is the largest dataset for natural image text recognition currently available. Each word image has an associated 50 word lexicon (IIIT5k-50) and 1k word lexicon (IIIT5k-1k).

{\bf IIIT Scene Text Retrieval (STR)~\cite{Mishra13}}
is a text based image retrieval dataset also collected with Google image search. Each of the 50 query words has an associated list of 10-50 images that contain the query word. There are also a large number of distractor images with no text downloaded from Flickr. In total there are 10k images and word bounding box annotation is not provided.

{\bf The IIIT Sports-10k dataset~\cite{Mishra13}}
is another text based image retrieval dataset constructed from frames of sports video. The images are low resolution and often noisy or blurred, with text generally located on advertisements and signboards, making this a challenging retrieval task. 10 query words are provided with 10k total images, without word bounding box annotations.

{\bf BBC News}
is a proprietary dataset of frames from the British Broadcasting Corporation (BBC) programmes that were broadcast between 2007 and 2012. Around 5000 hours of video (approximately 12 million frames) were processed to select 2.3 million keyframes at $1024 \times 768$ resolution. The videos are taken from a range of different BBC programmes on news and current affairs, including the BBC's Evening News programme. Text is often present in the frames from artificially inserted labels, subtitles, news-ticker text, and general scene text. No labels or annotations are provided for this dataset.

\begin{table*}[t]
\begin{center}\footnotesize
\begin{tabular}{|l|l|c|c|} \hline
\multicolumn{4}{|c|}{\bf Text Recognition Datasets}\\
\hline
Label & {\centering\footnotesize Description} & Lex. size & \#~images\\
\hline\hline
Synth        & {Our synthetically generated test dataset.} & 90k & 900k\\
\hline
IC03        & {ICDAR 2003~\cite{icdar2003dataset} test dataset.} & -- & 860\\
\hline
IC03-50     & {ICDAR 2003~\cite{icdar2003dataset} test dataset with fixed lexicon.} & 50 & 860\\
\hline
IC03-Full   & {ICDAR 2003~\cite{icdar2003dataset} test dataset with fixed lexicon.} & 860 & 860\\
\hline
SVT         & {SVT~\cite{Wang10} test dataset.} & -- & 647\\
\hline
SVT-50      & {SVT~\cite{Wang10} test dataset with fixed lexicon.} & 50 & 647\\
\hline
IC13      & {ICDAR 2013~\cite{ICDAR2013} test dataset.} & - & 1015\\
\hline
IIIT5k-50         & {IIIT5k~\cite{Mishra12} test dataset with fixed lexicon.} & 50& 3000\\
\hline
IIIT5k-1k      & {IIIT5k~\cite{Mishra12} test dataset with fixed lexicon.} & 1000 & 3000\\
\hline
\end{tabular}
\end{center}\vspace{-1em}
\caption{\small A description of the various \emph{text recognition} datasets evaluated on.}
\label{table:evalrecog}
\end{table*}

\begin{table*}[t]
\begin{center}\footnotesize
\begin{tabular}{|l|l|c|c|} \hline
\multicolumn{4}{|c|}{\bf Text Spotting Datasets}\\
\hline
Label & {\centering\footnotesize Description} & Lex. size & \#~images\\
\hline\hline
IC03        & {ICDAR 2003~\cite{icdar2003dataset} test dataset.} & -- & 251\\
\hline
IC03-50     & {ICDAR 2003~\cite{icdar2003dataset} test dataset with fixed lexicon.} & 50 & 251\\
\hline
IC03-Full   & {ICDAR 2003~\cite{icdar2003dataset} test dataset with fixed lexicon.} & 860 & 251\\
\hline
SVT         & {SVT~\cite{Wang10} test dataset.} & -- & 249\\
\hline
SVT-50      & {SVT~\cite{Wang10} test dataset with fixed lexicon.} & 50 & 249\\
\hline
IC11      & {ICDAR 2011~\cite{ICDAR11} test dataset.} & - & 255\\
\hline
IC13      & {ICDAR 2013~\cite{ICDAR2013} test dataset.} & - & 233\\
\hline
\end{tabular}
\end{center}\vspace{-1em}
\caption{\small A description of the various \emph{text spotting} datasets evaluated on.}
\label{table:evalspotting}
\end{table*}

\begin{table*}[t]
\begin{center}\footnotesize
\begin{tabular}{|l|l|c|c|} \hline
\multicolumn{4}{|c|}{\bf Text Retrieval Datasets}\\
\hline
Label & {\centering\footnotesize Description} & \#~queries & \#~images\\
\hline\hline
IC11        & {ICDAR 2011~\cite{ICDAR11} test dataset.} & 538 & 255\\
\hline
SVT         & {SVT~\cite{Wang10} test dataset.} & 427 & 249\\
\hline
STR    & {IIIT STR~\cite{Mishra13} text retrieval dataset.} & 50 & 10k\\
\hline
Sports    & {IIIT Sports-10k~\cite{Mishra13} text retrieval dataset.} & 10 & 10k\\
\hline
BBC News    & {A dataset of keyframes from BBC News video.} & - & 2.3m\\
\hline
\end{tabular}
\end{center}\vspace{-1em}
\caption{\small A description of the various \emph{text retrieval} datasets evaluated on.}
\label{table:evalretrieval}
\end{table*}

\subsection{Implementation Details}
\label{sec:implementation}

\begin{figure}
\begin{center}
\includegraphics[width=0.8\linewidth]{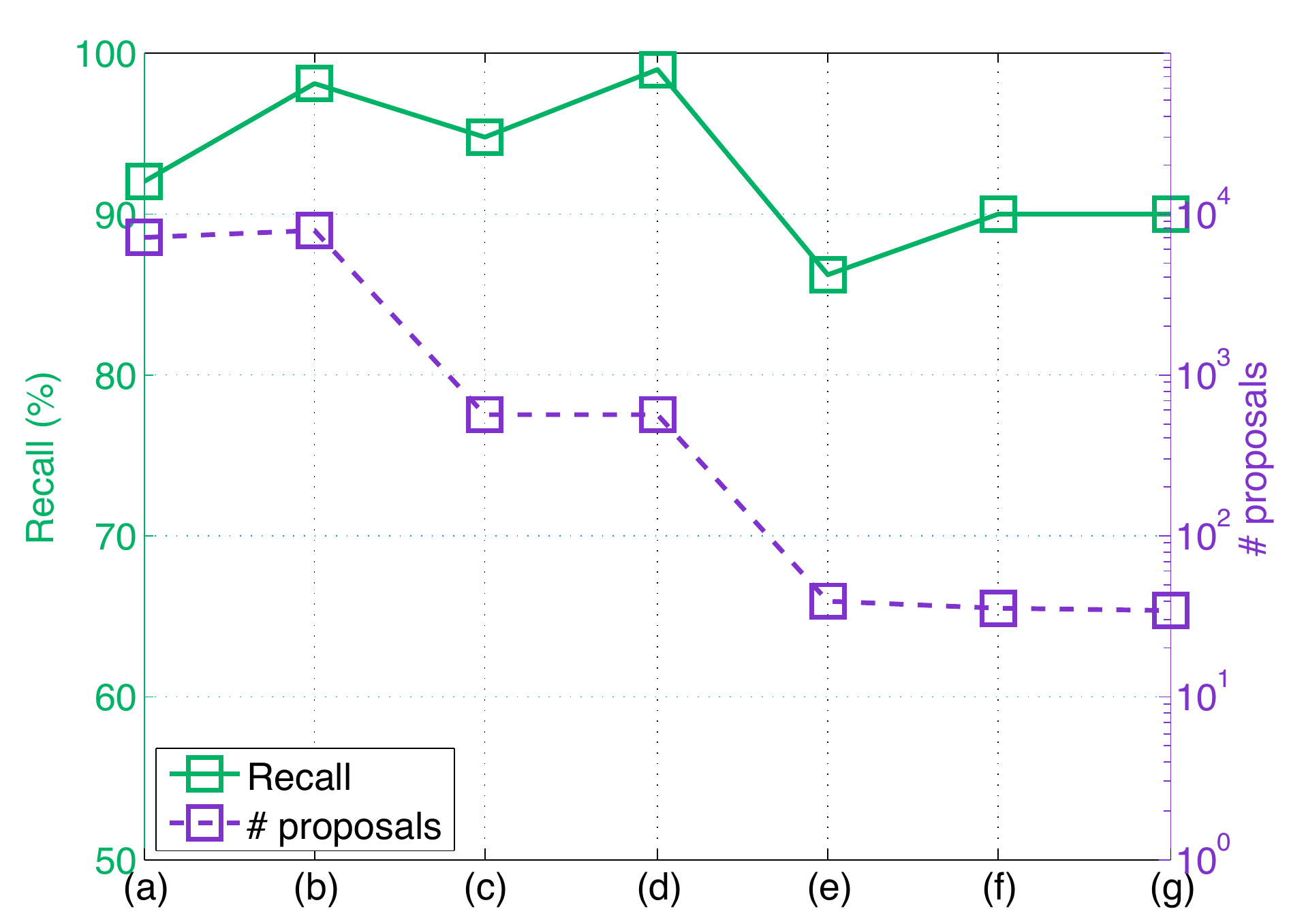} 
\caption{The recall and the average number of proposals per image at each stage of the pipeline on IC03. (a) Edge Box proposals, (b) ACF detector proposals, (c) Proposal filtering, (d) Bounding box regression, (e) Regression NMS round 1, (f) Regression NMS round 2, (g) Regression NMS round 3. The recall computed is detection recall across the dataset (\ie ignoring the recognition label) at 0.5 overlap.}
\label{fig:recall}
\end{center}
\end{figure}

We train a single model for each of the stages in our pipeline, and hyper parameters are selected using training datasets of ICDAR and SVT. Exactly the same pipeline, with the same models and hyper parameters are used for all datasets and experiments. This highlights the generalisability of our end-to-end framework to different datasets and tasks. The progression of detection recall and the number of proposals as the pipeline progresses can be seen in Fig.~\ref{fig:recall}.

\subsubsection{Edge Boxes \& ACF Detector}

The Edge Box detector has a number of hyper parameters, controlling the stride of evaluation and non maximal suppression. We use the default values of $\alpha=0.65$ and $\beta=0.75$ (see~\cite{Zitnick14} for details of these parameters). In practice, we saw little effect of changing these parameters in combined recall.

For the ACF detector, we set the number of decision trees to be 32, 128, 512 for each round of bootstrapping. For feature aggregation, we use $4 \times 4$ blocks smoothed with $[1~2~1]/4$ filter, with 8 scales per octave. As the detector is trained for a particular aspect ratio, we perform detection at multiple aspect ratios in the range $[1,1.2,1.4,\ldots,3]$ to account for variable sized words. We train on 30k cropped $32 \times 100$ positive word samples amalgamated from a number of training datasets as outlined in~\cite{Jaderberg14a}, and randomly sample negative patches from 11k images which do not contain text.

\begin{figure}
\begin{center}
\includegraphics[width=\linewidth]{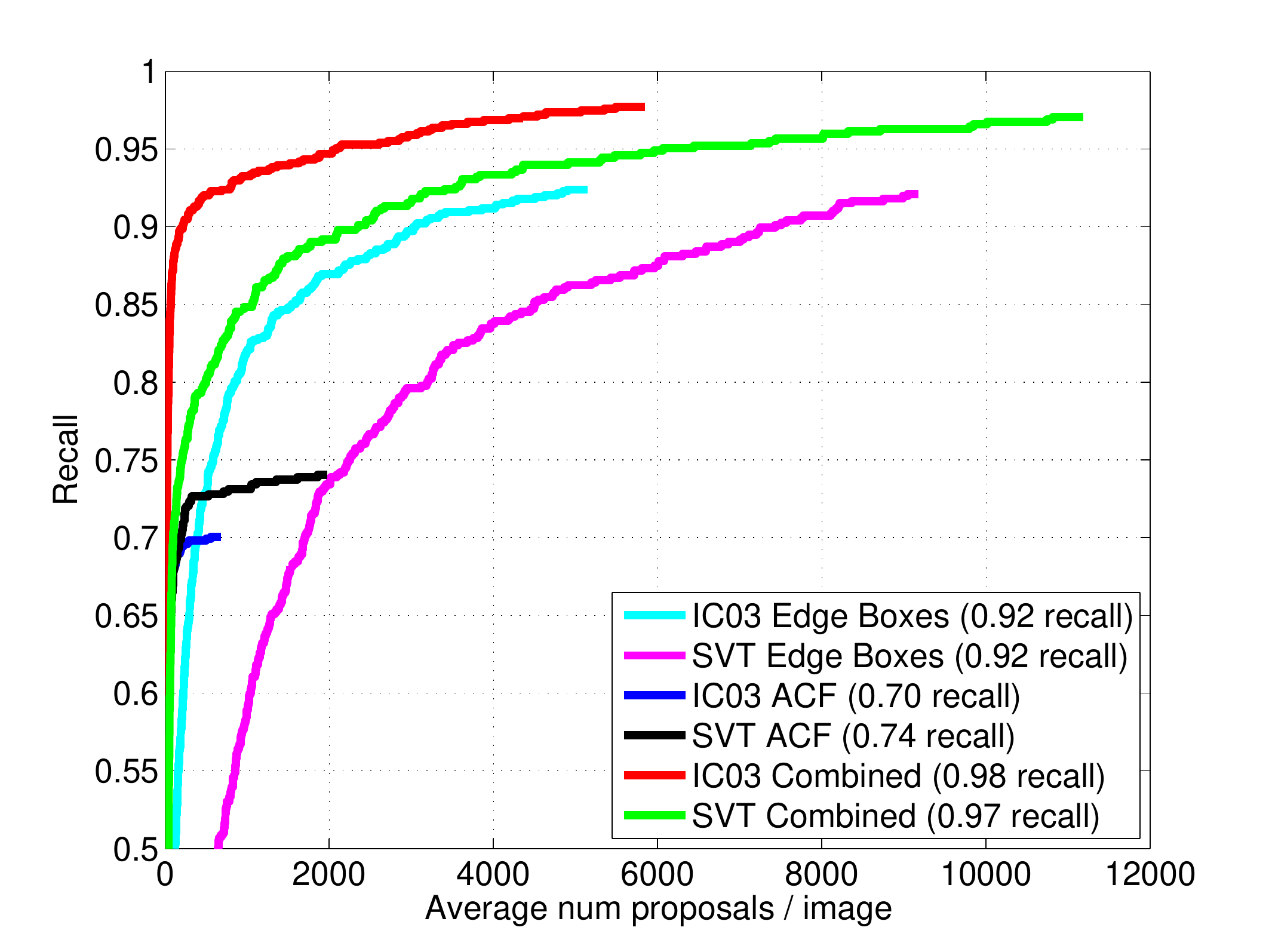} 
\caption{The 0.5 overlap recall of different region proposal algorithms. The recall displayed in the legend for each method gives the maximum recall achieved. The curves are generated by decreasing the minimum score for a proposal to be valid, and terminate when no more proposals can be found.}
\label{fig:ebrecall}
\end{center}
\end{figure}

Fig.~\ref{fig:ebrecall} shows the performance of our proposal generation stage. The recall at 0.5 overlap of groundtruth labelled words in the IC03 and SVT datasets is shown as a function of the number of proposal regions generated per image. The maximum recall achieved using Edge Boxes is 92\%, and the maximum recall achieved by the ACF detector is around 70\%. However, combining the proposals from each method increases the recall to 98\% at 6k proposals and 97\% at 11k proposals for IC03 and SVT respectively. The average maximum overlap of a particular proposal with a groundtruth bounding box is 0.82 on IC03 and 0.77 on SVT, suggesting the region proposal techniques produce some accurate detections amongst the thousands of false-positives. 

This high recall and high overlap gives a good starting point to the rest of our pipeline, and has greatly reduced the search space of word detections from the tens of millions of possible bounding boxes to around 10k proposals per image.

\subsubsection{Random Forest Word Classifier}
The random forest word/no-word binary classifier acts on cropped region proposals. These are resampled to a fixed $32 \times 100$ size, and HOG features extracted with a cell size of 4, resulting in $h \in \mathbb{R}^{8 \times 25 \times 36}$, a 7200-dimensional descriptor. The random forest classifier consists of 10 trees with a maximum depth of 64.

For training, region proposals are extracted as we describe in Section~\ref{sec:proposals} on the training datasets of ICDAR and SVT, with positive bounding box samples defined as having at least 0.5 overlap with groundtruth, and negative samples as less than 0.3 with groundtruth. Due to the abundance of negative samples, we randomly sample an equal number of negative samples to positive samples, giving 300k positive and 400k negative training samples.

Once trained, the result is a very effective false-positive filter. We select an operating probability threshold of 0.5, giving 96.6\% and 94.8\% recall on IC03 and SVT positive proposal regions respectively. This filtering reduces the total number of region proposals to on average 650 (IC03) and 900 (SVT) proposals per image. 

\subsubsection{Bounding Box Regressor}
The bounding box regression CNN consists of four convolutional layers with stride 1 with \\$\{filter~size,~number~of~filters\}$ of $\{5,64\}$, $\{5,128\}$, $\{3,256\}$, $\{3,512\}$ for each layer from input respectively, followed by two fully-connected layers with 4k units and 4 units (one for each regression variable). All hidden layers are followed by rectified linear non-linearities, the inputs to convolutional layers are zero-padded to preserve dimensionality, and the convolutional layers are followed by $2 \times 2$ max pooling. The fixed sized input to the CNN is a $32 \times 100$ greyscale image which is zero centred by subtracting the image mean and normalised by dividing by the standard deviation.

The CNN is trained with stochastic gradient descent (SGD) with dropout~\cite{Hinton12} on the fully-connected layers to reduce overfitting, minimising the $L_2$ distance between the estimated and groundtruth bounding boxes (Equation~\ref{eqn:regcost}). We used 700k training examples of bounding box proposals with greater than 0.5 overlap with groundtruth computed on the ICDAR and SVT training datasets.

Before the regression, the average positive proposal region (with over 0.5 overlap with groundtruth) had an overlap of 0.61 and 0.60 on IC03 and SVT. The CNN improves this average positive overlap to 0.88 and 0.70 for IC03 and SVT.

\subsubsection{Text Recognition CNN}
The text recognition CNN consists of eight weight layers -- five convolutional layers and three fully-connected layers. The convolutional layers have the following \\$\{filter~size,~ number~of~filters\}$: $\{5,64\}$, $\{5,128\}$, \\$\{3,256\}$, $\{3,512\}$, $\{3,512\}$. The first two fully-connected layers have 4k units and the final fully-connected layer has the same number of units as as number of words in the dictionary -- 90k words in our case. The final classification layer is followed by a softmax normalisation layer. Rectified linear non-linearities follow every hidden layer, and all but the fourth convolutional layers are followed by $2 \times 2$ max pooling. The inputs to convolutional layers are zero-padded to preserve dimensionality. The fixed sized input to the CNN is a $32 \times 100$ greyscale image which is zero centred by subtracting the image mean and normalised by dividing by the standard deviation. 

We train the network on Synth training data, back-propagating the standard multinomial logistic regression loss. Optimisation uses SGD with dropout regularisation of fully-connected layers, and we dynamically lower the learning rate as training progresses. With uniform sampling of classes in training data, we found the SGD batch size must be at least a fifth of the total number of classes in order for the network to train. 

For very large numbers of classes (\ie~over 5k classes), the SGD batch size required to train effectively becomes large, slowing down training a lot. Therefore, for large dictionaries, we perform \emph{incremental training} to avoid requiring a prohibitively large batch size. This involves initially training the network with 5k classes until partial convergence, after which an extra 5k classes are added. The original weights are copied for the previously trained classes, with the extra classification layer weights being randomly initialised. The network is then allowed to continue training, with the extra randomly initialised weights and classes causing a spike in training error, which is quickly trained away. This process of allowing partial convergence on a subset of the classes, before adding in more classes, is repeated until the full number of desired classes is reached.

At evaluation-time we do not do any data augmentation. If a lexicon is provided, we set the language prior $P(w|\mathcal{L})$ to be equal probability for lexicon words, otherwise zero. In the absence of a lexicon, $P(w|\mathcal{L})$ is calculated as the frequency of word $w$ in a corpus (we use the opensubtitles.org English corpus) with power law normalisation. In total, this model contains around 500 million parameters and can process a word in 2.2ms on a GPU with a custom version of Caffe~\cite{Jia13}.

\paragraph{Recognition Results.}
We evaluate the accuracy of our text recognition model over a wide range of datasets and lexicon sizes. We follow the standard evaluation protocol by~\cite{Wang11} and perform recognition on the words containing only alphanumeric characters and at least three characters.

The results are shown in Table~\ref{table:recognition}, and highlight the exceptional performance of our deep CNN. Although we train on purely synthetic data, with no human annotation, our model obtains significant improvements on state-of-the-art accuracy across all standard datasets. On IC03-50, the recognition problem is largely solved with 98.7\% accuracy -- only 11 mistakes out of 860 test samples -- and we significantly outperform the previous state-of-the-art \cite{Bissacco13} on SVT-50 by 5\% and IC13 by 3\%. Compared to the ICDAR datasets, the SVT accuracy, without the constraints of a dataset-specific lexicon, is lower at 80.7\%. This reflects the difficulty of the SVT dataset as image samples can be of very low quality, noisy, and with low contrast. The Synth dataset accuracy shows that our model really can recognise word samples consistently across the whole 90k dictionary.

\begin{figure}
\begin{center}
\includegraphics[width=0.8\linewidth]{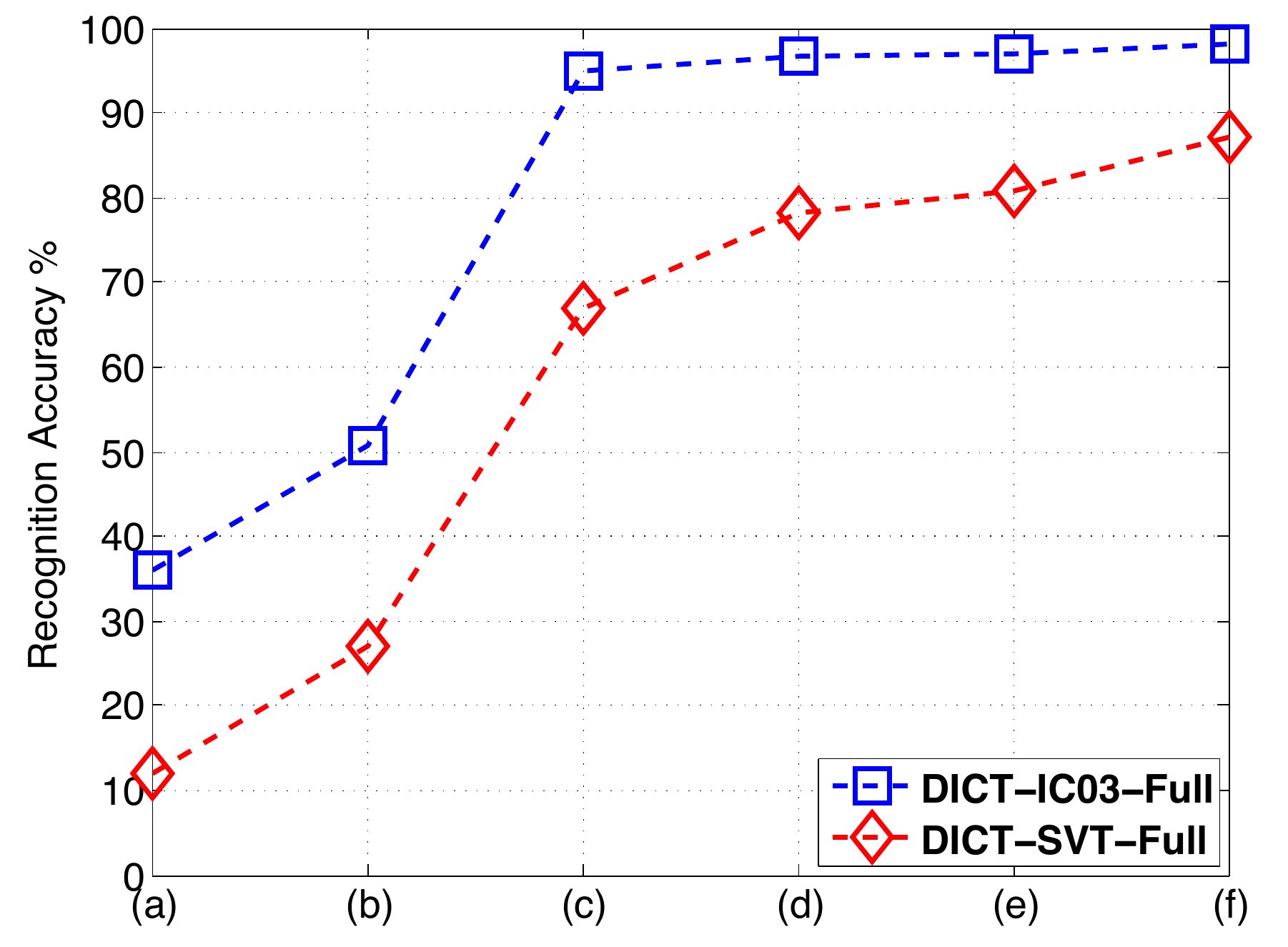} 
\caption{The recognition accuracies of text recognition models trained on just the IC03 lexicon (DICT-03-Full) and just the SVT lexicon (DICT-SVT-Full), evaluated on IC03 and SVT respectively. The models are trained on purely synthetic data with increasing levels of sophistication of the synthetic data. (a) Black text rendered on a white background with a single font, Droid Sans. (b) Incorporating all of Google fonts. (c) Adding background, foreground, and border colouring. (d) Adding perspective distortions. (e) Adding noise, blur and elastic distortions. (f) Adding natural image blending -- this gives an additional 6.2\% accuracy on SVT. The final accuracies on IC03 and SVT are 98.1\% and 87.0\% respectively.}
\label{fig:synthprogress}
\end{center}
\end{figure}

\paragraph{Synthetic Data Effects.}
As an additional experiment, we look into the contribution that the various stages of the synthetic data generation engine in Section~\ref{sec:synthdata} make to real-world recognition accuracy. We define two reduced recognition models (for speed of computation) with dictionaries covering just the IC03 and SVT full lexicons, denoted as DICT-IC03-Full and DICT-SVT-Full respectively, which are tested only on their respective datasets. We repeatedly train these models from scratch, with the same training procedure, but with increasing levels of sophistication of synthetic data. Fig.~\ref{fig:synthprogress} shows how the test accuracy of these models increases as more sophisticated synthetic training data is used. The addition of random image-layer colouring causes a notable increase in performance (+44\% on IC03 and +40\% on SVT), as does the addition of natural image blending (+1\% on IC03 and +6\% on SVT). It is interesting to observe a much larger increase in accuracy through incorporating natural image blending on the SVT dataset compared to the IC03 dataset. This is most likely due to the fact that there are more varied and complex backgrounds to text in SVT compared to in IC03.

\setlength{\tabcolsep}{3pt}
\begin{table*}[t]
\begin{center}
\begin{tabular}[t]{|l||c|c|c|c|c|c|c|c|c|}
\cline{2-10} 
\multicolumn{1}{c|}{\centering ~} & 
\multicolumn{9}{|c|}{\bf Cropped Word Recognition Accuracy (\%)}\\
\hline
\multicolumn{1}{|c||}{\centering Model} & 
\multicolumn{1}{c|}{\centering Synth} &
\multicolumn{1}{c|}{\centering IC03-50} &
\multicolumn{1}{c|}{\centering IC03-Full} &
\multicolumn{1}{c|}{\centering IC03} &
\multicolumn{1}{c|}{\centering SVT-50} &
\multicolumn{1}{c|}{\centering SVT} &
\multicolumn{1}{c|}{\centering IC13}&
\multicolumn{1}{c|}{\centering IIIT5k-50}&
\multicolumn{1}{c|}{\centering IIIT5k-1k}\\
\hline\hline
\emph{Baseline (ABBYY)}~\cite{Wang11,Yao14} & - & 56.0 & 55.0 & - & 35.0 & - & - & 24.3 & -\\
\rowcolor{Gray}
Wang~\cite{Wang11}          & - & 76.0 & 62.0 & - & 57.0 & - & - & - & -\\
Mishra~\cite{Mishra12}      & - & 81.8 & 67.8 & - & 73.2 & - & - & 64.1 & 57.5\\
\rowcolor{Gray}
Novikova~\cite{Novikova12}  & - & 82.8 & - & - & 72.9 & - & - & - & -\\
Wang~\&~Wu~\cite{Wang12}    & - & 90.0 & 84.0 & - & 70.0 & - & - & - & -\\
\rowcolor{Gray}
Goel~\cite{Goel13}          & - & 89.7 & - & - & 77.3 & - & - & - & -\\
PhotoOCR~\cite{Bissacco13} & - & - & - & - & 90.4 & 78.0 & 87.6 & - & -\\
\rowcolor{Gray}
Alsharif~\cite{Alsharif13}  & - & 93.1 & 88.6 & 85.1\textsuperscript{*} & 74.3 & - & - & - & -\\
Almazan~\cite{Almazan14} & - & - & - & - & 89.2 & - & - & 91.2 & 82.1\\
\rowcolor{Gray}
Yao~\cite{Yao14}  & - & 88.5 & 80.3 & - & 75.9 & - & - & 80.2 & 69.3\\
Jaderberg~\cite{Jaderberg14a}  & - & 96.2 & 91.5 & - & 86.1 & - & - & - & -\\
\rowcolor{Gray}
Gordo~\cite{Gordo14} & - & - & - & - & 90.7 & - & - & 93.3 & 86.6\\
\hline
Proposed     & \bf 95.2 & \bf 98.7 & \bf 98.6 & \bf{93.3} & \bf 95.4 & \bf 80.7 & \bf 90.8 & \bf 97.1 & \bf 92.7\\
\hline
\end{tabular}
\end{center}
\vspace*{-1em}
\caption{\small Comparison to previous methods for text recognition accuracy -- where the groundtruth cropped word image is given as input. The ICDAR 2013 results given are case-insensitive. Bold results outperform previous state-of-the-art methods. The baseline method is from a commercially available document OCR system. \textsuperscript{*}Recognition is constrained to a dictionary of 50k words.}
\label{table:recognition}
\end{table*}

\subsection{Text Spotting}
\label{sec:textspotting}

\setlength{\tabcolsep}{3pt}
\begin{table*}[t]
\begin{center}
\begin{tabular}[t]{|l||c|c|c|c|c|c|c|c|c|}
\cline{2-10} 
\multicolumn{1}{c|}{\centering ~} & 
\multicolumn{9}{|c|}{\bf End-to-End Text Spotting (F-measure \%)}\\
\hline
\multicolumn{1}{|c||}{\centering Model} & 
\multicolumn{1}{c|}{\centering IC03-50} &
\multicolumn{1}{c|}{\centering IC03-Full} &
\multicolumn{1}{c|}{\centering IC03} &
\multicolumn{1}{c|}{\centering IC03\textsuperscript{+}} &
\multicolumn{1}{c|}{\centering SVT-50} &
\multicolumn{1}{c|}{\centering SVT} &
\multicolumn{1}{c|}{\centering IC11} &
\multicolumn{1}{c|}{\centering IC11\textsuperscript{+}} &
\multicolumn{1}{c|}{\centering IC13}\\
\hline\hline
Neumann~\cite{Neumann11} & - & - & - & 41 & - & - & - & - & -\\
\rowcolor{Gray}
Wang~\cite{Wang11} & 68 & 51 & - & - & 38 & - & - & - & -\\
Wang \& Wu~\cite{Wang12} & 72 & 67 & - & - & 46 & - & - & - & -\\
\rowcolor{Gray}
Neumann~\cite{Neumann13} & - & - & - & - & - & - & - & 45 & -\\
Alsharif~\cite{Alsharif13}  & 77 & 70 & 63\textsuperscript{*} & - & 48 & - & - & - & -\\
\rowcolor{Gray}
Jaderberg~\cite{Jaderberg14a}  & 80 & 75 & - & - & 56 & - & - & - & -\\
\hline
Proposed  & \bf 90 & \bf 86 & \bf 78 & \bf 72 & \bf 76 & \bf 53 & \bf 76 & \bf 69 & \bf 76\\
\rowcolor{Gray}
Proposed (0.3 IoU) & \bf 91 & \bf 87 & \bf 79 & \bf 73 & \bf 82 & \bf 57 & \bf 77 & \bf 70 & \bf 77\\
\hline
\end{tabular}
\end{center}
\vspace*{-1em}
\caption{\small Comparison to previous methods for end-to-end text spotting. Bold results outperform previous state-of-the-art methods. \textsuperscript{*}Recognition is constrained to a dictionary of 50k words. \textsuperscript{+}Evaluation protocol described in~\cite{Neumann13}.}
\label{table:spotting}
\end{table*}

\begin{figure*}
\begin{center}
\begin{tabular}{ccc}
\includegraphics[width=0.33\linewidth]{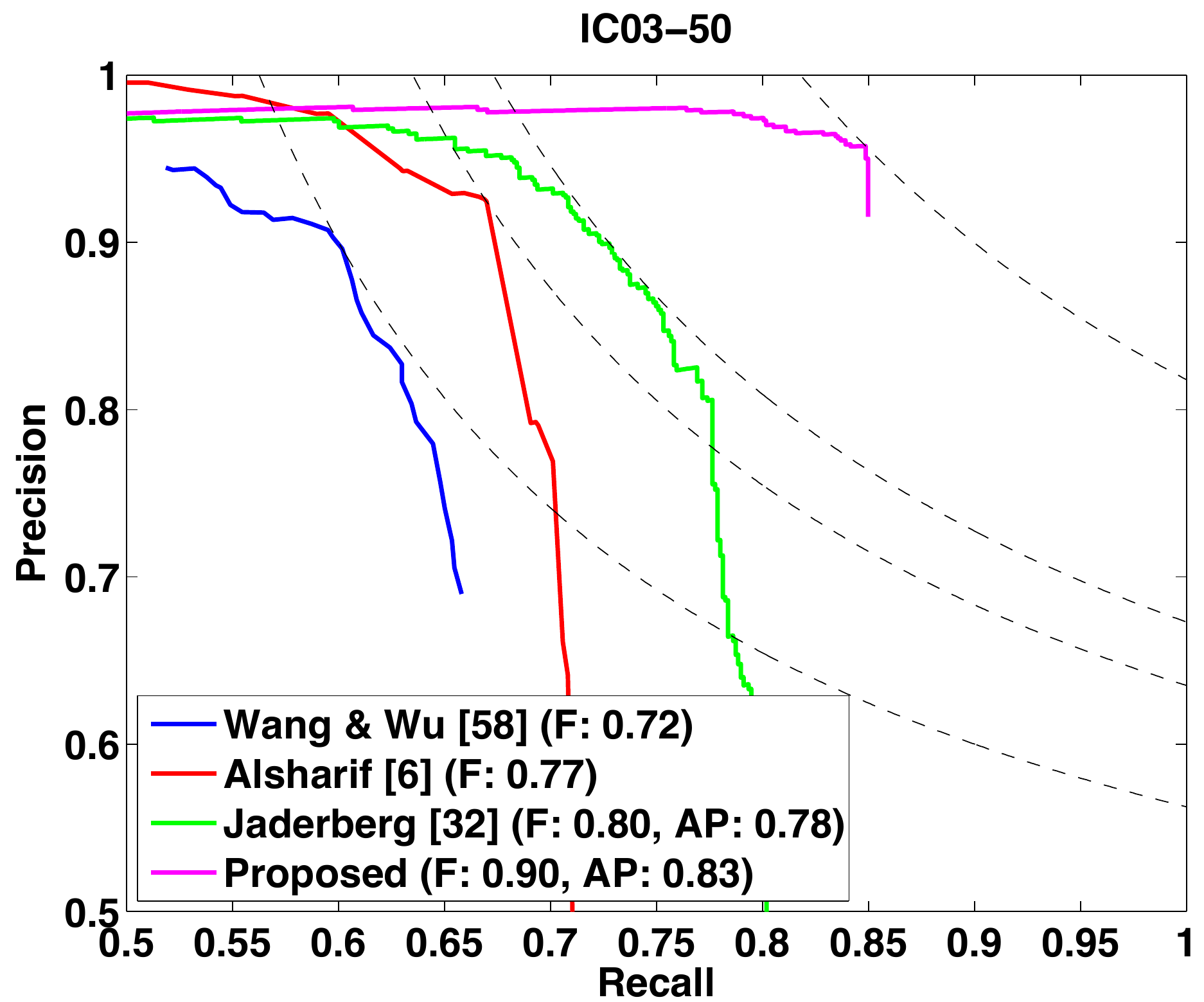}&
\includegraphics[width=0.33\linewidth]{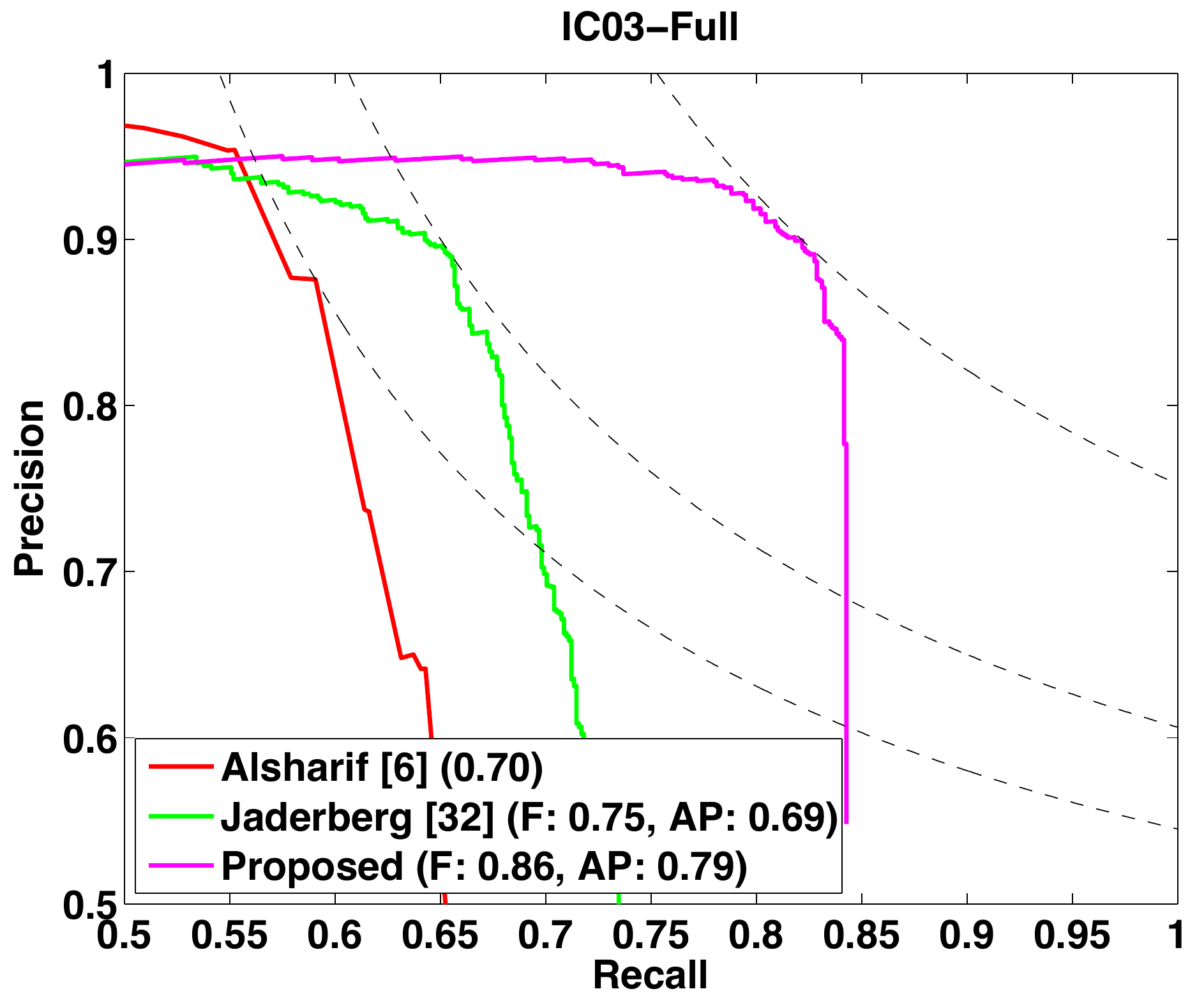}&
\includegraphics[width=0.33\linewidth]{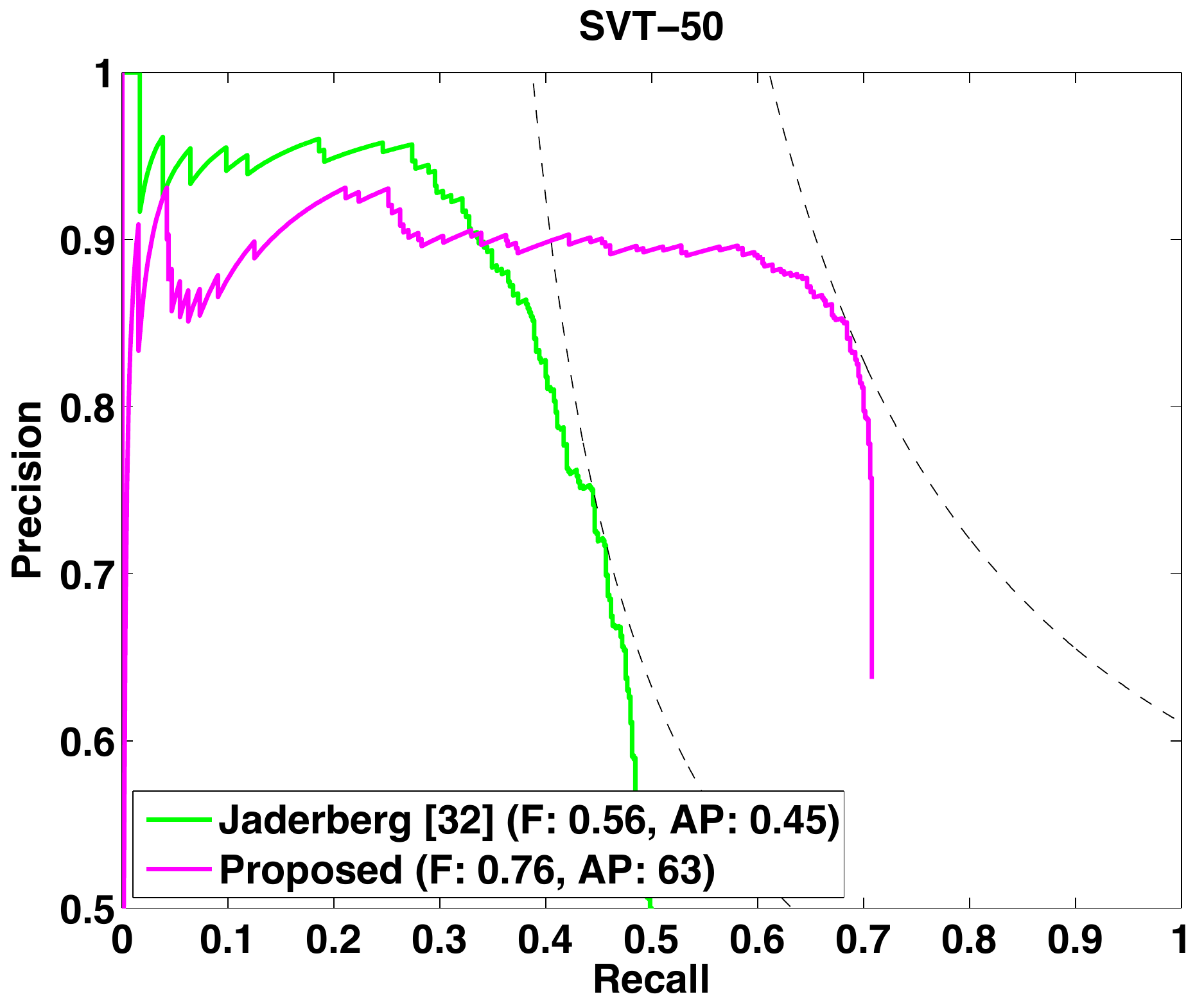}\\
(a) & (b) & (c)
\end{tabular}
\caption{The precision/recall curves on (a) the IC03-50 dataset, (b) the IC03-Full dataset, and (c) the SVT-50 dataset. The lines of constant F-measure are shown at the maximum F-measure point of each curve. The results from \cite{Wang12,Alsharif13} were extracted from the papers.}
\label{fig:prcurves}
\end{center}
\end{figure*}

\setlength{\tabcolsep}{3pt}
\begin{table*}[t]
\begin{center}
\begin{tabular}[t]{|l||c|c|c|c|c|c|}
\cline{2-7} 
\multicolumn{1}{c|}{\centering ~} & 
\multicolumn{6}{|c|}{\bf Text Based Image Retrieval}\\
\hline
\multicolumn{1}{|c||}{\centering Model} & 
\multicolumn{1}{c|}{\centering IC11 (mAP)} &
\multicolumn{1}{c|}{\centering SVT (mAP)} &
\multicolumn{1}{c|}{\centering STR (mAP)} &
\multicolumn{1}{c|}{\centering Sports (mAP)} &
\multicolumn{1}{c|}{\centering Sports (P@10)}&
\multicolumn{1}{c|}{\centering Sports (P@20)}\\
\hline\hline
Wang~\cite{Wang11}\textsuperscript{*}  & - & 21.3 & - & - & - & -\\
\rowcolor{Gray}
Neumann~\cite{Neumann12}\textsuperscript{*}  & - & 23.3 & - & - & - & -\\
Mishra~\cite{Mishra13}  & 65.3 & 56.2 & 42.7 & - & 44.8 & 43.4\\
\hline
\rowcolor{Gray}
Proposed  & \bf 90.3 & \bf 86.3 & \bf 66.5 & \bf 66.1 & \bf 91.0 & \bf 92.5\\
\hline
\end{tabular}
\end{center}
\vspace*{-1em}
\caption{\small Comparison to previous methods for text based image retrieval. We report mean average precision (mAP) for IC11, SVT, STR, and Sports, and also report top-$n$ retrieval to compute precision at $n$ (P@$n$) on Sports. Bold results outperform previous state-of-the-art methods. \textsuperscript{*}Experiments were performed by Mishra \etal in \cite{Mishra13}, not by the original authors.}
\label{table:retrieval}
\end{table*}

\begin{figure*}
\begin{center}
\begin{tabular}{ccc}
\includegraphics[width=0.33\linewidth]{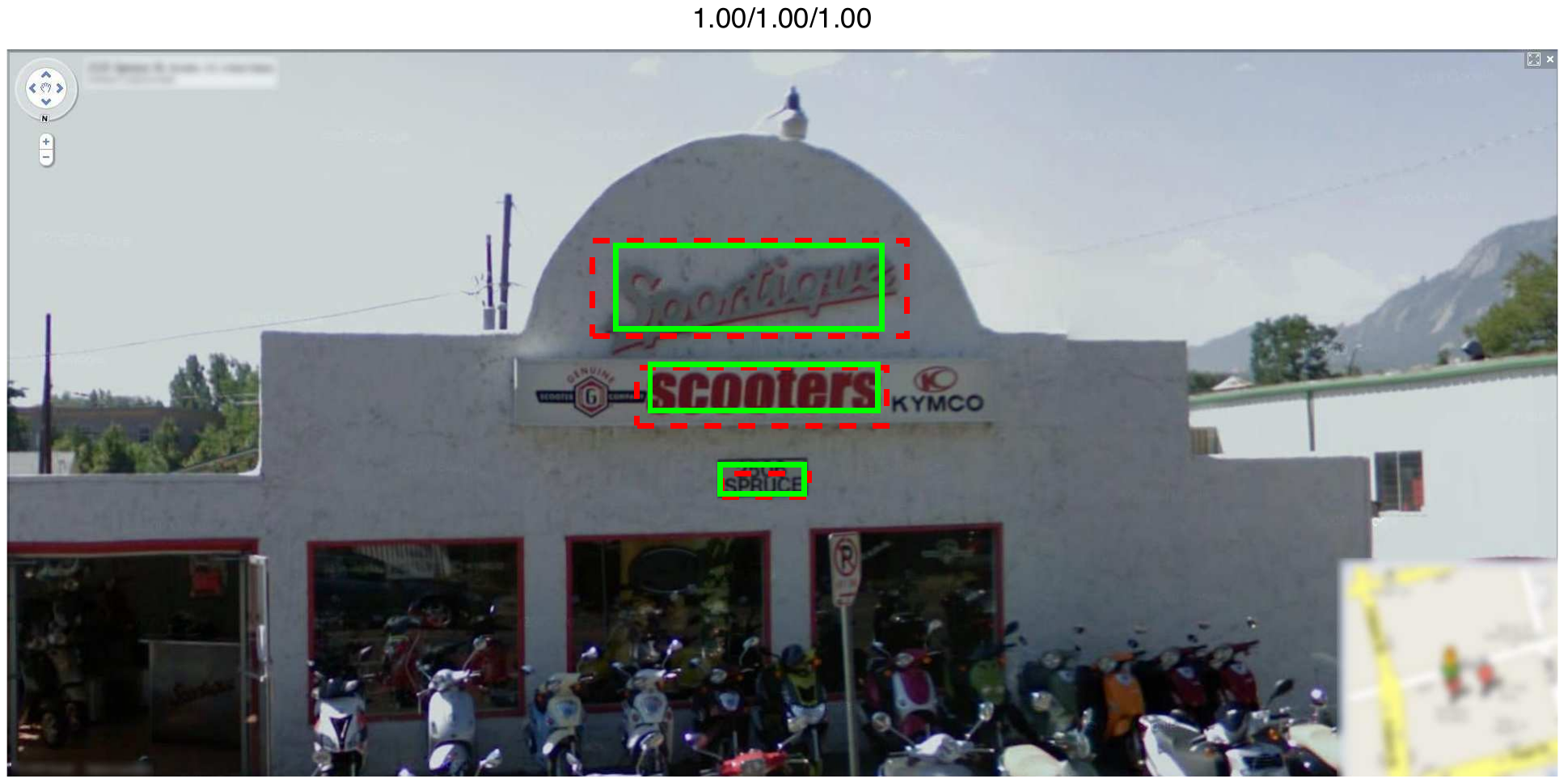}&
\includegraphics[width=0.33\linewidth]{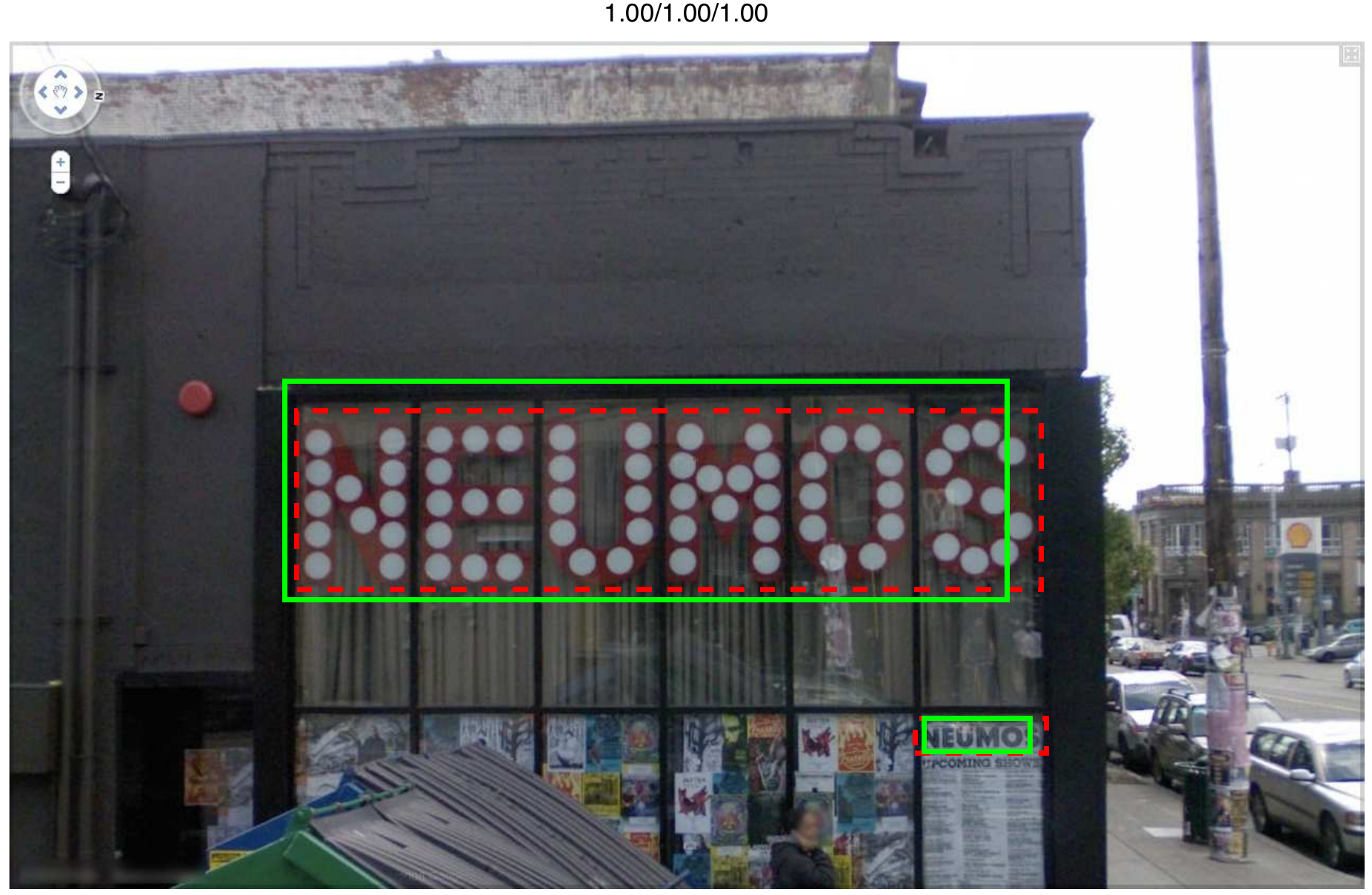}&
\includegraphics[width=0.33\linewidth]{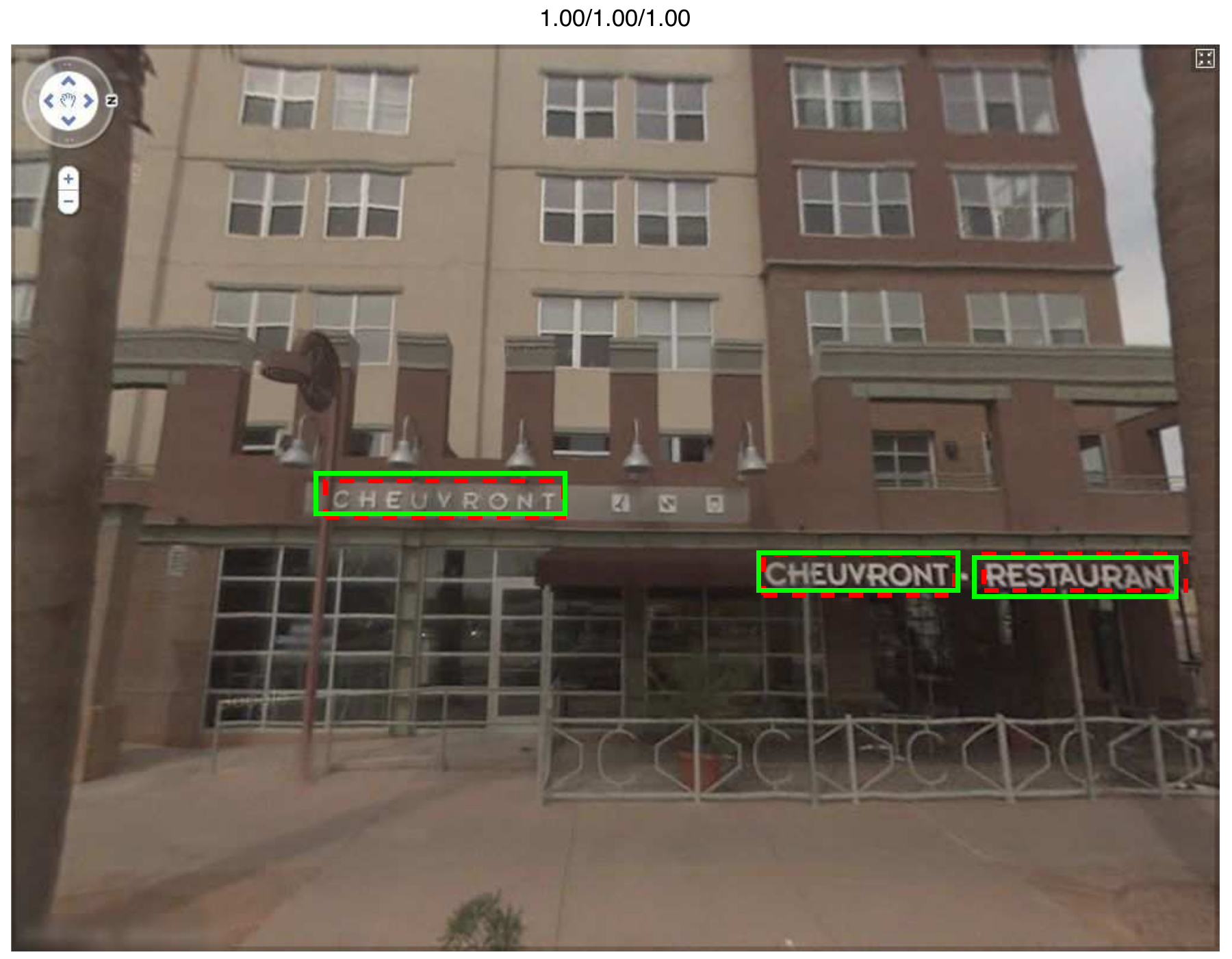}\\
\includegraphics[width=0.33\linewidth]{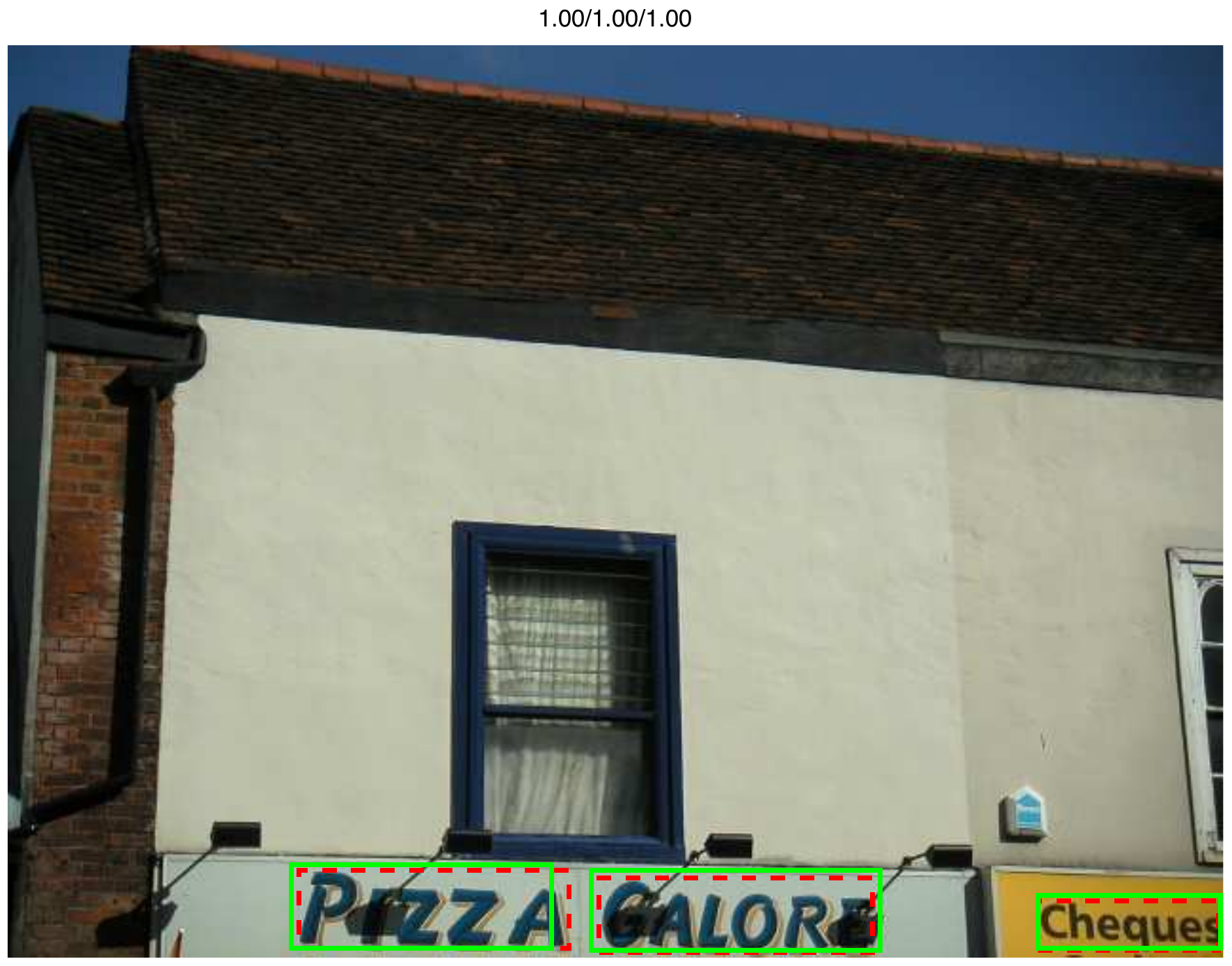}&
\includegraphics[width=0.33\linewidth]{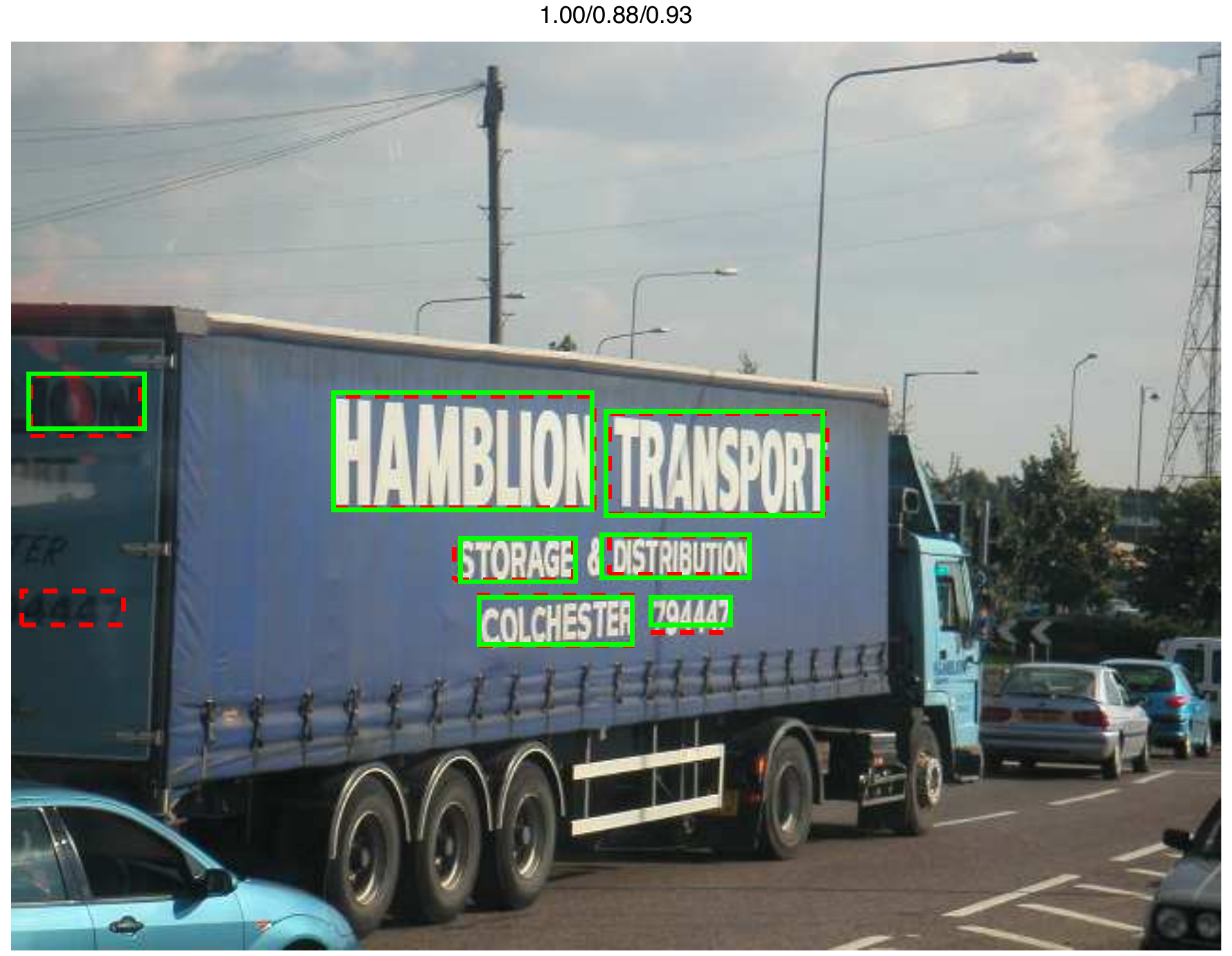}&
\includegraphics[width=0.33\linewidth]{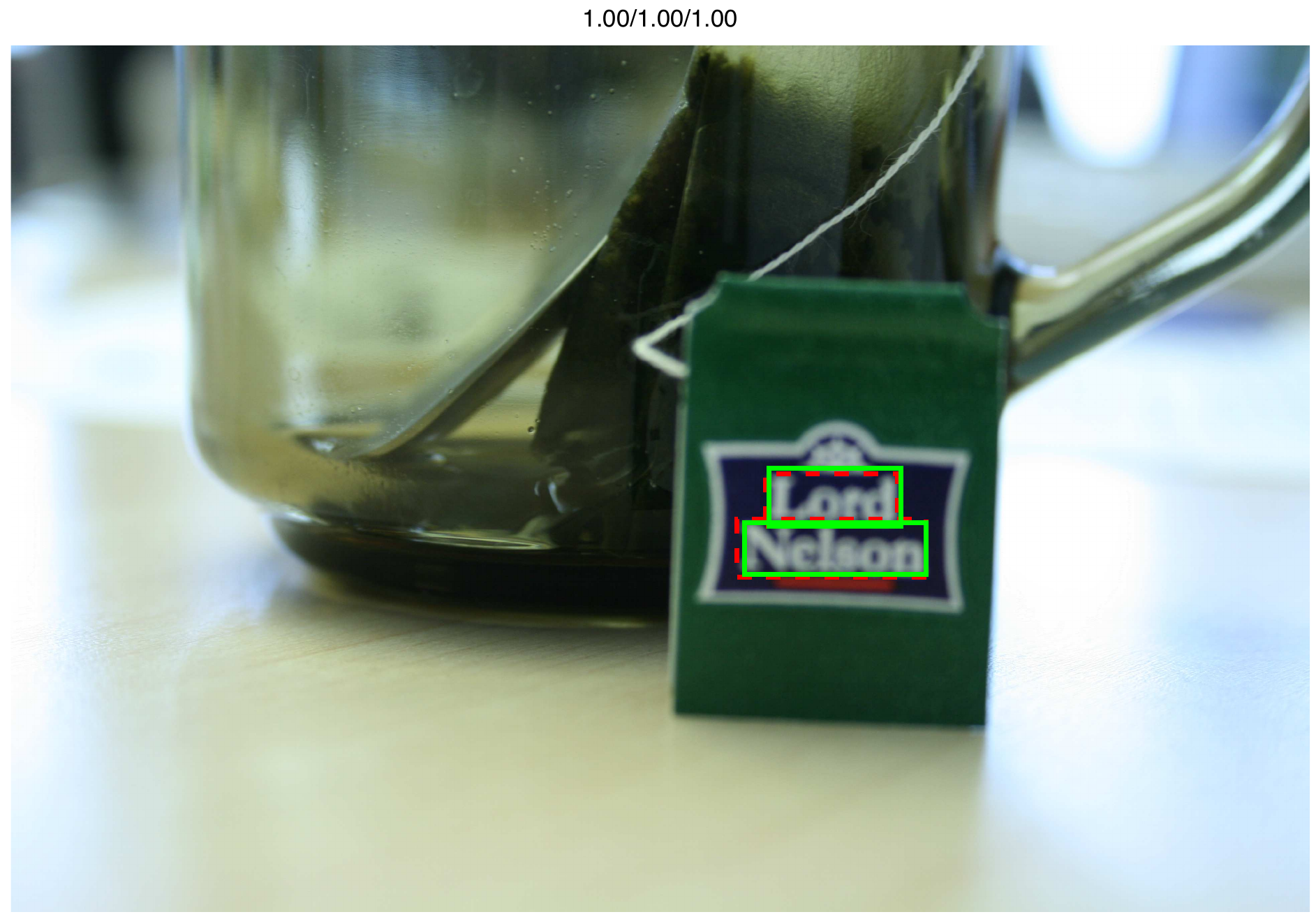}\\
\end{tabular}
\caption{Some example text spotting results from SVT-50 (top row) and IC11 (bottom row). Red dashed shows groundtruth and green shows correctly localised and recognised results. P/R/F figures are given above each image.}
\label{fig:examples}
\end{center}
\end{figure*}

In the text spotting task, the goal is to localise and recognise the words in the test images. Unless otherwise stated, we follow the standard evaluation protocol by \cite{Wang11} and ignore all words that contain alphanumeric characters and are not at least three characters long. A positive recognition result is only valid if the detection bounding box has at least 0.5 overlap (IoU) with the groundtruth.

Table~\ref{table:spotting} shows the results of our text spotting pipeline compared to previous methods. We report the global F-measure over all images in the dataset. Across all datasets, our pipeline drastically outperforms all previous methods. On SVT-50, we increase the state-of-the-art by +20\% to a P/R/F (precision/recall/F-measure) of 0.85/0.68/0.76 compared to 0.73/0.45/0.56 in \cite{Jaderberg14a}. Similarly impressive improvements can be seen on IC03, where in all lexicon scenarios we improve F-measure by at least +10\%, reaching a P/R/F of 0.96/0.85/0.90. Looking at the precision/recall curves in Fig.~\ref{fig:prcurves}, we can see that our pipeline manages to maintain very high recall, and the recognition score of our text recognition system is a strong cue to the suitability of a detection.

We also give results across all datasets when no lexicon is given.  As expected, the F-measure suffers from the lack of lexicon constraints, though is still significantly higher than other comparable work. It should be noted that the SVT dataset is only partially annotated. This means that the precision (and therefore F-measure) is much lower than the true precision if fully annotated, since many words that are detected are not annotated and are therefore recorded as false-positives. We can however report recall on SVT-50 and SVT of 71\% and 59\% respectively.

Interestingly, when the overlap threshold is reduced to 0.3 (last row of Table~\ref{table:spotting}), we see a small improvement across ICDAR datasets and a large +8\% improvement on SVT-50. This implies that our text recognition CNN is able to accurately recognise even loosely cropped detections. Ignoring the requirement of correctly recognising the words, \ie performing purely word detection, we get an F-measure of 0.85 and 0.81 for IC03 and IC11.

Some example text spotting results are shown in Fig.~\ref{fig:examples}. Since our pipeline does not rely on connected component based algorithms or explicit character recognition, we can detect and recognise disjoint, occluded and blurry words. 

A common failure mode of our system is the missing of words due to the lack of suitable proposal regions, especially apparent for slanted or vertical text, something which is not explicitly modelled in our framework. Also the detection of sub-words or multiple words together can cause false-positive results.

\subsection{Image Retrieval}
\label{sec:retrieval}

\begin{figure*}
\centering
\begin{center}
\begin{tabular}{c|c|c}
\texttt{hollywood} -- P@100: 100\% & \texttt{boris johnson} -- P@100: 100\% & \texttt{vision} -- P@100: 93\%\\
\includegraphics[width=0.3\linewidth]{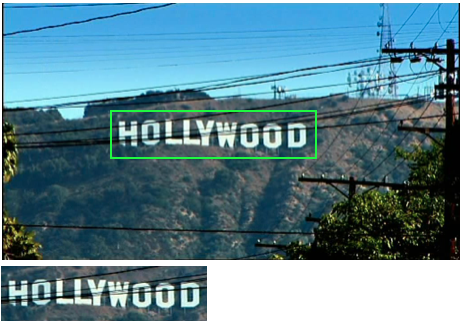} & 
\includegraphics[width=0.3\linewidth]{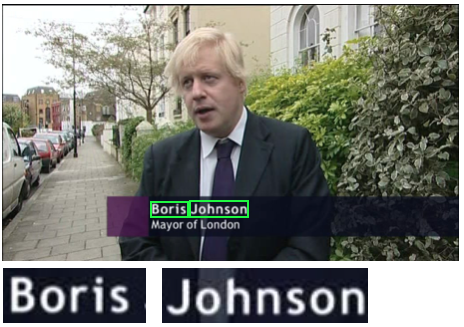} &
\includegraphics[width=0.3\linewidth]{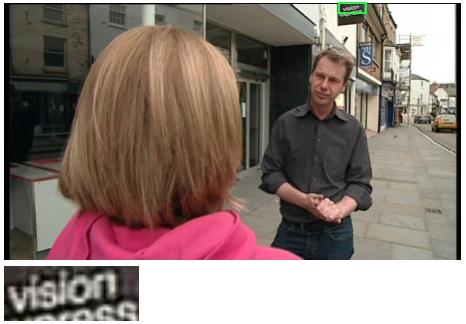}\\
\includegraphics[width=0.3\linewidth]{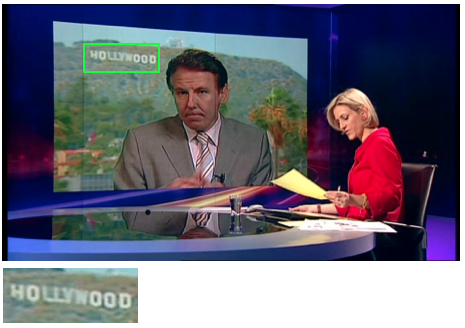} &
\includegraphics[width=0.3\linewidth]{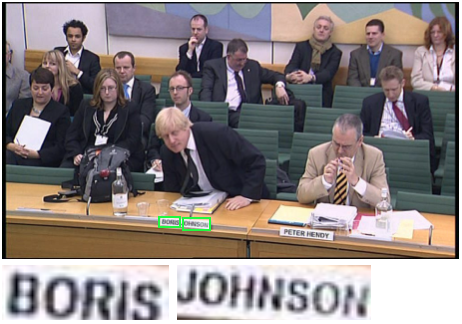} &
\includegraphics[width=0.3\linewidth]{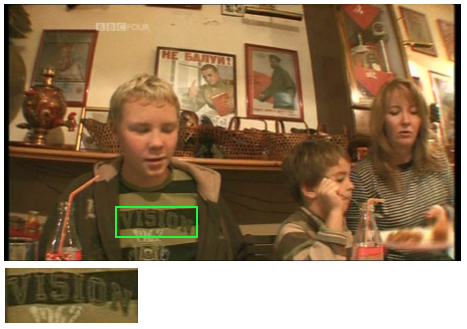}\\
\end{tabular}
\caption{The top two retrieval results for three queries on our BBC News dataset -- \texttt{hollywood}, \texttt{boris johnson}, and \texttt{vision}. The frames and associated videos are retrieved from 5k hours of BBC video. We give the precision at 100 (P@100) for these queries, equivalent to the first page of results of our web application.}
\label{fig:bbc}
\end{center}
\end{figure*}

We also apply our pipeline to the task of text based image retrieval. Given a text query, the images containing the query text must be returned. 

This task is evaluated using the framework of \cite{Mishra13}, with the results shown in Table~\ref{table:retrieval}. For each defined query, we retrieve a ranked list of all the images of the dataset and compute the average precision (AP) for each query, reporting the mean average precision (mAP) over all queries. We significantly outperform Mishra \etal across all datasets -- we obtain an mAP on IC11 of 90.3\%, compared to 65.3\% from~\cite{Mishra13}. Our method scales seamlessly to the larger Sports dataset, where our system achieves a precision at 20 images (P@20) of 92.5\%, more than doubling that of 43.4\% from \cite{Mishra13}. Mishra~\etal~\cite{Mishra13} also report retrieval results on SVT for released implementations of other text spotting algorithms. The method from Wang~\etal~\cite{Wang11} achieves 21.3\% mAP, the method from Neumann~\etal \cite{Neumann12} acheives 23.3\% mAP and the method proposed by \cite{Mishra13} itself achieves 56.2\% mAP, compared to our own result of 86.3\% mAP.

\begin{figure*}
\begin{center}
\begin{tabular}{c|c}
\bf SVT: \texttt{apartments} & \bf Sports: \texttt{castrol}\\
1. Score: 0.219 -- Groundtruth: 0 & 1. Score: 1.0 -- Groundtruth: 1\\
\includegraphics[width=0.4\linewidth]{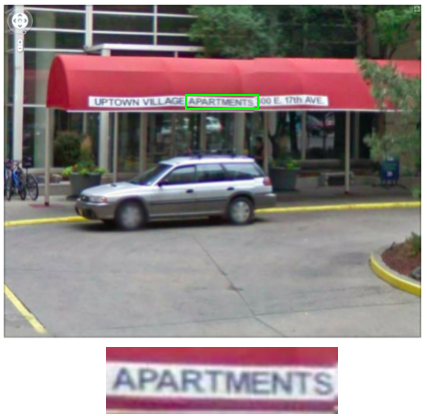}&\includegraphics[width=0.4\linewidth]{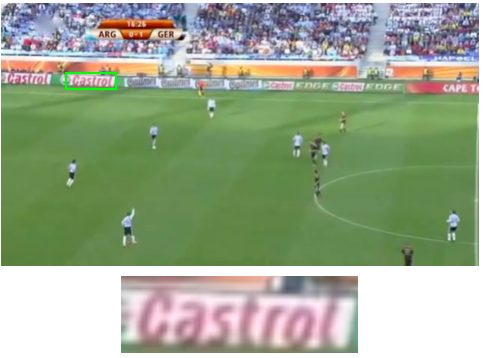}\\
\\
\\
2. Score: 0.022 -- Groundtruth: 1 & 2. Score: 1.0 -- Groundtruth: 0\\
\includegraphics[width=0.4\linewidth]{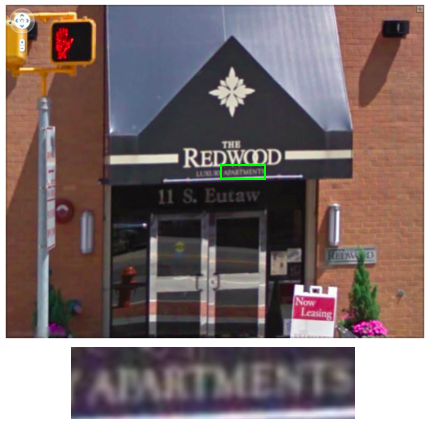}&\includegraphics[width=0.4\linewidth]{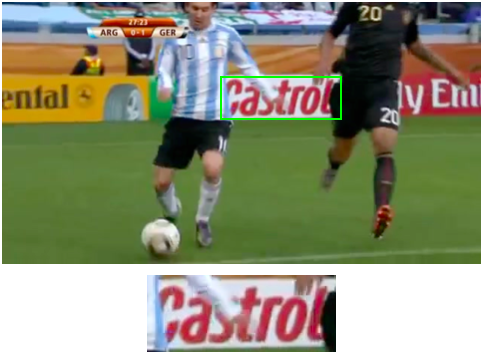}
\end{tabular}
\caption{An illustration of the problems with incomplete annotations in test datasets. We show examples of the top two results for the query \texttt{apartments} on the SVT dataset and the query \texttt{castrol} on the Sports dataset. All retrieved images contain the query word (green box), but some of the results have incomplete annotations, and so although the query word is present the result is labelled as incorrect. This leads to a reported AP of 0.5 instead of 1.0 for the SVT query, and a reported P@2 of 0.5 instead of the true P@2 of 1.0.}
\label{fig:retrievalmistake}
\end{center}
\end{figure*}

However, as with the text spotting results for SVT, our retrieval results suffer from incomplete annotations on SVT and Sports datasets -- Fig.~\ref{fig:retrievalmistake} shows how precision is hurt by this problem. The consequence is that the true mAP on SVT is higher than the reported mAP of 86.3\%. 

Depending on the image resolution, our algorithm takes approximately 5-20s to compute the end-to-end results per image on a single CPU core and single GPU. We analyse the time taken for each stage of our pipeline on the SVT dataset, which has an average image size of $1260 \times 860$, showing the results in Table~\ref{table:timings}. Since we reduce the number of proposals throughout the pipeline, we can allow the processing time per proposal to increase while keeping the total processing time for each stage stable. This affords us the use of more computationally complex features and classifiers as the pipeline progresses. Our method can be trivially parallelised, meaning we can process 1-2 images per second on a high-performance workstation with 16 physical CPU cores and 4 commodity GPUs. 


\begin{table}
\begin{center}
\footnotesize
\begin{tabular}{|l|c|c|c|} \hline
Stage & \#~proposals & Time & Time/proposal\\
\hline\hline
(a) Edge Boxes & $> 10^7$ & 2.2s & $<0.002$ms\\
(b) ACF detector & $> 10^7$ & 2.1s & $<0.002$ms\\
(c) RF filter & $10^4$ & 1.8s & 0.18ms\\
(d) CNN regression & $10^3$ & 1.2s & 1.2ms\\
(e) CNN recognition & $10^3$ & 2.2s & 2.2ms\\
\hline
\end{tabular}
\caption{The processing time for each stage of the pipeline evaluated on the SVT dataset on a single CPU core and single GPU. As the pipeline progresses from (a)-(e), the number of proposals is reduced (starting from all possible bounding boxes), allowing us to increase our computational budget per proposal while keeping the overall processing time for each stage comparable.}
\label{table:timings}
\end{center}
\end{table}

The high precision and speed of our pipeline allows us to process huge datasets for practical search applications. We demonstrate this on a 5000 hour BBC News dataset. Building a search engine and front-end web application around our image retrieval pipeline allows a user to instantly search for visual occurrences of text within the huge video dataset. This works exceptionally well, with Fig.~\ref{fig:bbc} showing some example retrieval results from our visual search engine. While we do not have groundtruth annotations to quantify the retrieval performance on this dataset, we measure the precision at 100 (P@100) for the test queries in Fig.~\ref{fig:bbc}, showing a P@100 of 100\% for the queries \texttt{hollywood} and \texttt{boris johnson}, and 93\% for \texttt{vision}. These results demonstrate the scalable nature of our framework.

\section{Conclusions}
\label{sec:conclusions}
In this work we have presented an end-to-end text reading pipeline -- a detection and recognition system for text in natural scene images. This general system works remarkably well for both text spotting and image retrieval tasks, significantly improving the performance on both tasks over all previous methods, on all standard datasets, without any dataset-specific tuning. This is largely down to a very high recall proposal stage and a text recognition model that achieves superior accuracy to all previous systems. Our system is fast and scalable -- we demonstrate seamless scalability from datasets of hundreds of images to being able to process datasets of millions of images for instant text based image retrieval without any perceivable degradation in accuracy. Additionally, the ability of our recognition model to be trained purely on synthetic data allows our system to be easily re-trained for recognition of other languages or scripts, without any human labelling effort. 

We set a new benchmark for text spotting and image retrieval. Moving into the future, we hope to explore additional recognition models to allow the recognition of unknown words and arbitrary strings.

\section*{Acknowledgements}
This work was supported by the EPSRC and ERC grant VisRec no. 228180. We gratefully acknowledge the support of NVIDIA Corporation with the donation of the GPUs used for this research. We thank the BBC and in particular Rob Cooper for access to data and video processing resources.

\bibliographystyle{spmpsci}

\small\bibliography{longstrings,vgg_local,vgg_other,max_bib,mybib}

\end{document}